%% file: 0_main.tex
\documentclass[sigconf]{acmart}
\AtBeginDocument{%
  \providecommand\BibTeX{{%
    \normalfont B\kern-0.5em{\scshape i\kern-0.25em b}\kern-0.8em\TeX}}}

\copyrightyear{2021}
\acmYear{2021}
\setcopyright{acmcopyright}\acmConference[SIGIR '21]{Proceedings of the 44th International ACM SIGIR Conference on Research and Development in Information Retrieval}{July 11--15, 2021}{Virtual Event, Canada}
\acmBooktitle{Proceedings of the 44th International ACM SIGIR Conference on Research and Development in Information Retrieval (SIGIR '21), July 11--15, 2021, Virtual Event, Canada}
\acmPrice{15.00}
\acmDOI{10.1145/3404835.3462971}
\acmISBN{978-1-4503-8037-9/21/07}

\acmSubmissionID{1067}

\usepackage{booktabs} 
\usepackage{helvet}  
\usepackage{courier}  
\usepackage{url}  
\usepackage{graphicx}  
\usepackage{footnote}
\usepackage{subfigure}
\usepackage{amsmath}
\usepackage{amsfonts}
\usepackage{multirow}
\usepackage{bm}
\usepackage{enumitem}
\usepackage{color}

\usepackage{algorithm}
\usepackage{algpseudocode}
\floatname{algorithm}{Algorithm}

\renewcommand{\textrightarrow}{$\longrightarrow$}

\newcommand{\ie}{\emph{i.e., }}
\newcommand{\eg}{\emph{e.g., }}
\newcommand{\etal}{\emph{et al. }}

\newcommand{\wrt}{\emph{w.r.t. }}

\emergencystretch=\maxdimen
\usepackage{xparse}
\usepackage{color}
\NewDocumentCommand{\cyx}{ mO{} }{\textcolor{red}{\textsuperscript{\textit{Yixin}}\textsf{\textbf{\small[#1]}}}}

\begin{document}
\fancyhead{}
\title{Should Graph Convolution Trust Neighbors? A Simple Causal Inference Method}

\author{Fuli Feng$^{12*}$, Weiran Huang$^{3}$, Xiangnan He$^{4}$, Xin Xin$^5$, Qifan Wang$^6$, Tat-Seng Chua$^{2}$}

\def\authors{Fuli Feng, Weiran Huang, Xiangnan He, Xin Xin, Qifan Wang, Tat-Seng Chua}

\affiliation{
\institution{$^1$Sea-NExT Joint Lab, $^2$National University of Singapore, $^3$The Chinese University of Hong Kong\\$^4$University of Science and Technology of China, $^5$University of Glasgow, $^6$Google US}
\country{}
}

\email{fulifeng93@gmail.com,huangweiran1998@outlook.com,xiangnanhe@gmail.com}
\email{x.xin.1@research.gla.ac.uk,wqfcr@google.com,dcscts@nus.edu.sg}

\thanks{$*$Corresponding author. This research is supported by the Sea-NExT Joint Lab, National Natural Science Foundation of China (U19A2079) and National Key Research and Development Program of China (2020AAA0106000).}






\renewcommand{\shortauthors}{Feng and Huang, et al.}

\begin{abstract}
Graph Convolutional Network (GCN) is an emerging technique for information retrieval (IR) applications.
While GCN assumes the homophily property of a graph, 
real-world graphs are never perfect: the local structure of a node may contain discrepancy, \eg the labels of a node's neighbors could vary.
This pushes us to consider the discrepancy of local structure in GCN modeling. 
Existing work approaches this issue by introducing an additional module such as graph attention, which is expected to learn the contribution of each neighbor. 
However, such module may not work reliably as expected, 
especially when there lacks supervision signal, \eg when the labeled data is small. Moreover, existing methods focus on modeling the nodes in the training data, and never consider the local structure discrepancy of testing nodes. 


This work focuses on the local structure discrepancy issue for testing nodes, which has received little scrutiny. 
From a novel perspective of causality, we investigate whether a GCN should trust the local structure of a testing node when predicting its label.
To this end, we analyze the working mechanism of GCN with causal graph, estimating the \textit{causal effect} of a node's local structure for the prediction.
The idea is simple yet effective: given a trained GCN model, we first intervene the prediction by blocking the graph structure; we then compare the original prediction with the intervened prediction to assess the \textit{causal effect} of the local structure on the prediction. 
Through this way, we can eliminate the impact of local structure discrepancy and make more accurate prediction.
Extensive experiments on seven node classification datasets show that our method effectively enhances the inference stage of GCN.

\end{abstract}



\maketitle

\input{1_int}
\input{2_pre.tex}
\input{4_met}
\input{5_exp}
\input{3_pilot}
\input{6_rel}
\input{7_con}

\bibliographystyle{ACM-Reference-Format}
\bibliography{refers}

\appendix

\end{document}

%% file: 1_int.tex
\section{Introduction}
\begin{figure}[t]
	\centering
	\includegraphics[width=0.42\textwidth]{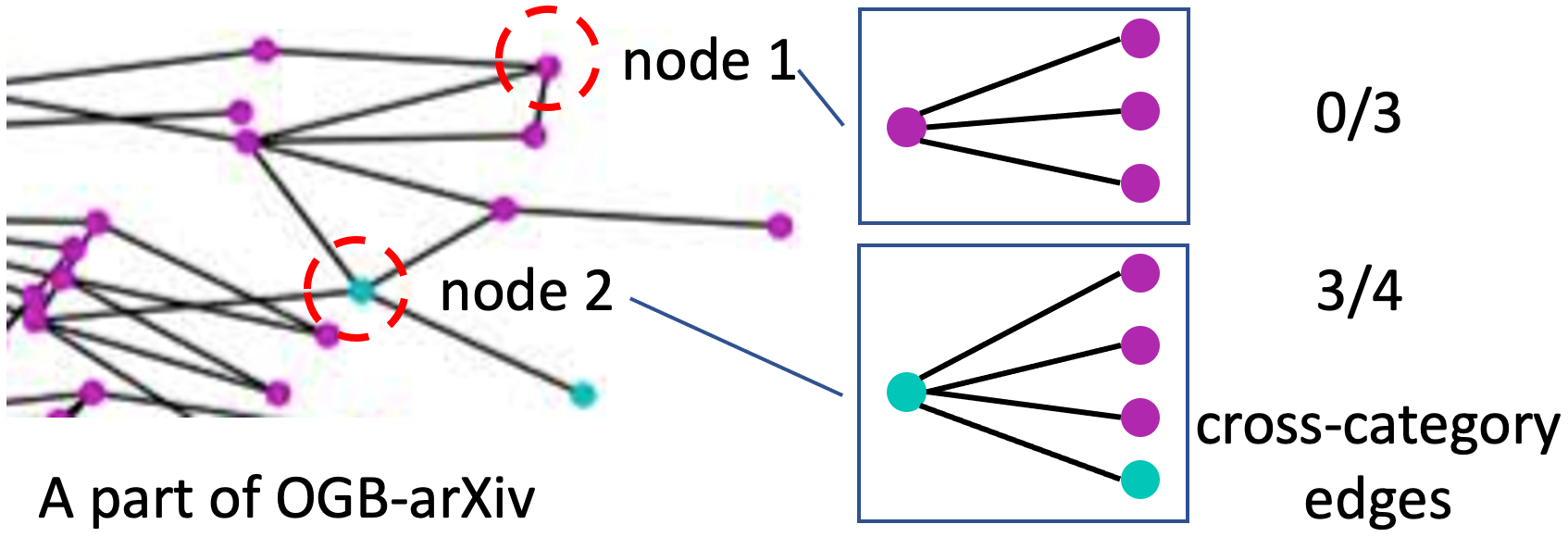}
	\vspace{-0.4cm}
	\caption{Illustration of local structure discrepancy in OGB-arXiv~\cite{OGB}, a citation graph of papers. Nodes in different colors belong to different categories.
	}
	\vspace{-0.5cm}
	\label{fig:cross}
\end{figure}

GCN is being increasingly used in IR applications, ranging from search engines~\cite{mao2020item,yang2020biomedical}, recommender systems~\cite{NGCF,chen2019semisupervised,LightGCN,chen2020bias} to question-answering systems~\cite{zhang2020answer,hu2020residual}.
Its main idea is to augment a node's representation by aggregating the representations of its neighbors. 
In practice, GCN could face the local structure discrepancy issue~\cite{chen2019measuring} since real-world graphs usually exhibit locally varying structure. That is, nodes can exhibit inconsistent distributions of local structure properties such as homophily and degree. 
Figure~\ref{fig:cross} shows an example in a document citation graph~\cite{OGB}, where the local structure centered at $node 1$ and $node 2$ has different properties regarding cross-category edges\footnote{This work focuses on the discrepancy \wrt cross-category connections.}. Undoubtedly, applying the same aggregation over $node 1$ and $node 2$ will lead to inferior node representations. Therefore, it is essential for GCN to account for the local structure discrepancy issue.

Existing work considers this issue by equipping GCN with an adaptive locality module~\cite{JKNet,GAT}, which learns to adjust the contribution of neighbors. 
Most of the efforts focus on the attention mechanism, such as neighbor attention~\cite{GAT} and hop attention~\cite{DAGNN}. 
Ideally, the attention weight could downweigh the neighbors that causes discrepancy, \eg the neighbors of different categories with the target node.
However, graph attention is not easy to be trained well in practice, especially for hard semi-supervised learning setting that has very limited labeled data~\cite{knyazev2019understanding}. 
Moreover, existing methods mainly consider the nodes in the training data, ignoring the local structure discrepancy on the testing nodes, which however are the decisive factor for model generalization. It is thus insufficient to resolve the discrepancy issue by adjusting the architecture of GCN.

In this work, we argue that it is essential to empower the inference stage of a trained GCN with the ability of handling the local structure discrepancy issue. 
In real-world applications, the graph structure typically evolves along time, resulting in structure discrepancy between the training data and testing data. Moreover, the testing node can be newly coming (\eg a new user), which may exhibit properties different from the training nodes~\cite{tang2020investigating}. 
However, the one-pass inference procedure of existing GCNs 
indiscriminately uses the learned model parameters to make prediction for all testing nodes, lacking the capability of handling the structure discrepancy. This work aims to bridge the gap by upgrading the GCN inference to be node-specific according to the extent of structure discrepancy.

To achieve the target, the key lies in analyzing the prediction generation process of GCN on each node and estimating to what extent accounting for node's neighbors affects its prediction, \ie the causal effect of the local structure on GCN prediction. 
According to the evidence that model output can reflect feature discrepancy~\cite{saito2018adversarial}, we have a key assumption that the GCN output provides evidence on the properties of the local structure centered at a testing node. 
For instance, if the local structure exhibits properties distinct from the seen ones, the model will be uncertain about its prediction when the neighbors are taken into account. 
Accordingly, we should downweigh the contribution of neighbors to reduce the impact of the discrepancy on the prediction. 
Inherently, both a node's features and neighbors are the causes of the prediction for the node. 
By distinguishing the two causal effects, we can assess revise the model prediction in a node-specific manner.

To this end, we resort to the language of causal graph~\cite{pearl2009causality} to describe the causal relations in GCN prediction. We propose a \textit{Causal GCN Inference} (CGI) model, which adjusts the prediction of a trained GCN according to the causal effect of the local structure. 
In particular, CGI first calls for causal intervention that blocks the graph structure and forces the GCN to user a node's own features to make prediction. CGI then makes choice between the intervened prediction and the original prediction, according to the causal effect of the local structure, prediction confidence, and other factors that characterize the prediction. 
Intuitively, CGI is expected to choose the intervened prediction (\ie trusting self) when facing a testing node with local structure discrepancy.
To learn a good choice-making strategy, we devise it as a separate classifier, which is learned based on the trained GCN.
We demonstrate CGI on APPNP~\cite{APPNP}, one of the state-of-the-art GCN models for semi-supervised node classification. Extensive experiments on seven datasets validate the effectiveness of our approach. The codes are released at: \url{https://github.com/fulifeng/CGI}.

The main contributions of this work are summarized as follows:
\begin{itemize}[leftmargin=*]
    \item We achieve adaptive locality during GCN inference and propose an CGI model that is model-agnostic.  
    \item We formulate the causal graph of GCN working mechanism, and the estimation of causal intervention and causal uncertainty based on the causal graph. 
    \item We conduct experiments on seven node classification datasets to demonstrate the rationality of the proposed methods.
\end{itemize}

%% file: 2_pre.tex
\section{Preliminaries}
\paragraph{Node classification.} We represent a graph with $N$ nodes as $G = (\bm{A}, \bm{X})$, \ie an adjacency matrix $\bm{A} \in \mathbb{R}^{N \times N}$ associated with a feature matrix $\bm{X}=[\bm{x}_1,\bm{x}_2,\cdots,\bm{x}_N]^{T}\in\mathbb{R}^{N \times D}$. $\bm{A}$ describes the connections between nodes where $A_{ij} = 1$ means there is an edge between node $i$ and $j$, otherwise $A_{ij} = 0$. $D$ is the dimension of the input node features.
Node classification is one of the most popular analytic tasks on graph data. In the general problem setting, the label of $M$ nodes are given $\bm{Y} = [\bm{y}_1,\bm{y}_2,\cdots,\bm{y}_N]^{T}\in\mathbb{R}^{N \times L}$, where $L$ is the number of node categories and $\bm{y}_1$ is a one-hot vector. The target is to learn a classifier from the labeled nodes, formally,
\begin{equation}\small
    \begin{aligned}
        f(\bm{x}, \mathcal{N}(\bm{x}) | \bm{\theta}), ~ \mathcal{N}(\bm{x}) = \{\bm{x}_n | \bm{A}_{in} = 1\},
    \end{aligned}
\end{equation}
where $\bm{\theta}$ denotes the parameter of the classifier and $\mathcal{N}(\bm{x})$ denotes the neighbor nodes of the target node $\bm{x}$. 
Without loss of generality, we index the labeled nodes and testing nodes in the range of $[1, M]$ and $(T, N]$, respectively.
There are four popular settings with minor differences regarding the observability of testing nodes during model training and the amount of labeled nodes. Specifically,
\begin{itemize}[leftmargin=*]
    \item \textit{Inductive Full-supervised Learning}: In this setting, testing nodes are not included in the graph used for model training and all training nodes are labeled. That is, $M = T$ and learning the classifier with $f(\bm{X}_{tr} | \bm{A}_{tr}, \bm{\theta})$ where $\bm{X}_{tr} \in \mathbb{R}^{M \times D}$ and $\bm{A}_{tr}$ denotes the features and the subgraph of the training nodes.
    \item \textit{Inductive Semi-supervised Learning}~\cite{GraphSage}: In many real-world applications such as text classification~\cite{linmei2019heterogeneous}, it is unaffordable to label all the observed nodes, \ie only a small portion of the training nodes are labeled (in fact, $M \ll T$).
    \item \textit{Transductive Full-supervised Learning}~\cite{OGB}: In some cases, the graph is relatively stable, \ie no new node occurs, where the whole node graph $\bm{X}$ and $\bm{A}$ are utilized for model training.
    \item \textit{Transductive Semi-supervised Learning}~\cite{kipf2016semi}: In this setting, the whole graph is available for model training while only a small portion of the training nodes are labeled.
\end{itemize}
It should be noted that we do not restrict our problem setting to be a specific one like most previous work on GCN, since we focus on the general inference model.

\begin{figure*}[t]
	\centering
	\includegraphics[width=0.7\textwidth]{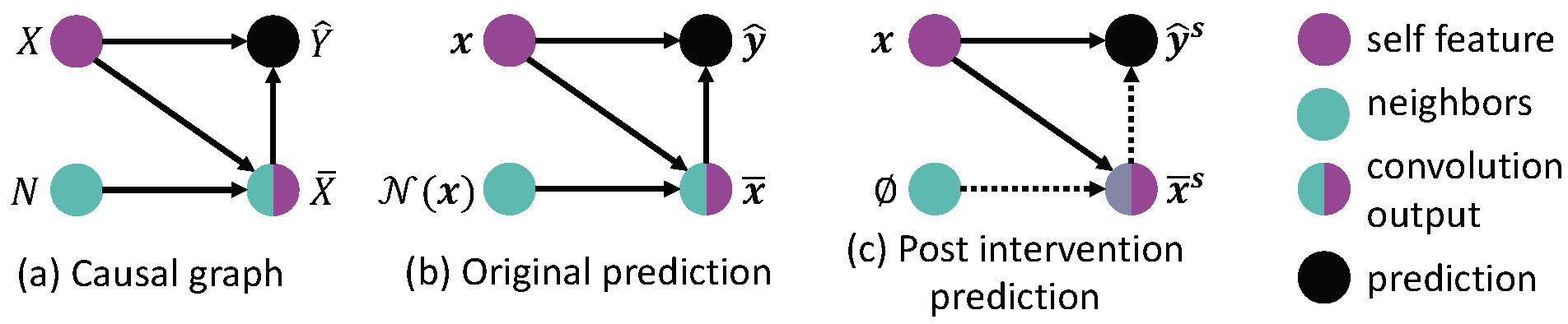}
	\vspace{-0.4cm}
	\caption{Cause-effect view of GCN. (a) Causal graph of GCN inference process; (b) making original prediction; (c) causal intervention $do(N = \emptyset)$ where dashed arrow means the effect from the predecessor is blocked.
	}
	\vspace{-0.3cm}
	\label{fig:causal_graph}
\end{figure*}

\paragraph{Graph Convolutional Network.} Taking the graph as input, GCN learns node representations to encode the graph structure and node features (the last layer makes predictions)~\cite{kipf2016semi}. 
The key operation of GCN is \textit{neighbor aggregation}, which can be abstracted as:
\begin{equation}\small
    \begin{aligned}
	\bar{\bm{x}} = AGG(\bm{x}, \{\bm{x}_{n} | \bm{x}_{n} \in \mathcal{N}(\bm{x})\}),
	\label{eq:gcn_layer}
    \end{aligned}
\end{equation}
where $AGG$ denotes the node aggregation operation such as a weighted summation~\cite{kipf2016semi}. $\bm{x}$ and $\bar{\bm{x}} \in \mathbb{R}^D$ are the origin representation of the target node (node features or representation at the previous layer) and the one after aggregating neighbor node features. Note that standard GCN layer typically consists a feature transformation, which is omitted for briefness. 

\paragraph{Adaptive locality.} In most GCN, the target node is equally treated as the neighbor nodes, \ie no additional operation except adding the edge for self-connection. Aiming to distinguish the contribution from target node and neighbor nodes, a self-weight $\alpha$ is utilized, \ie $\mathcal{N}(\alpha * \bm{x}, \{(1 - \alpha) * \bm{x}_{n} | n \in neighbor(\bm{x})\})$. 
More specifically, neighbor attention~\cite{GAT} is introduced to learn node specific weights, \ie $\mathcal{N}(\alpha \bm{x}, \{\alpha_{n} * \bm{x}_{n} | n \in neighbor(\bm{x})\})$. The weights $\alpha$ and $\alpha_n$ are calculated by an attention model such as multi-head attention~\cite{multi-head-attention} with the node representations $\bm{x}$ and $\bm{x}_{n}$ as inputs. Lastly, hop attention~\cite{DAGNN} is devised to adaptively aggregate the target node representations at different GCN layers $\bm{x}^{0}, \cdots, \bar{\bm{x}}^{k}, \cdots, \bar{\bm{x}}^{K}$ into a final representation. $\bar{\bm{x}}^{k}$ is the convolution output at the $k$-th layer which encodes the $k$-hop neighbors of the target node. For a target node that is expected to trust self more, the hop attention is expected to assign higher weight for $\bm{x}^{0}$. Most of these adaptive locality models are learned during model training except the self-weight $\alpha$ in GCN models like APPNP which is tuned upon the validation set.

\paragraph{Causal effect.} Causal effect is a concept in causal science~\cite{pearl2009causality}, which studies the influence among variables. Given two variables $X$ and $Y$, the causal effect of $X=x$ on $Y$ is to what extent changing the value of $X$ to $x$ affects the value of $Y$, which is abstracted as:
\begin{equation}\small
    \begin{aligned}
    Y_x - Y_{x^*},
    \end{aligned}
\end{equation}
where $Y_x$ and $Y_{x^*}$ are the outcomes of $Y$ with $X=x$ and $X=x^*$ as inputs, respectively. $x^*$ is the reference status of variable $X$, which is typically set as empty value (\eg zero) or the expectation of $X$.

%% file: 4_met.tex
\section{Methodology}
In this section, we first scrutinize the cause-effect factors in the inference procedure of GCN, and then introduce the proposed CGI. Assume that we are given a well-trained GCN $f(\bm{x}, \mathcal{N}(\bm{x}) | \hat{\theta})$, which is optimized over the training nodes according to the following objective function:
\begin{align}\label{eq:obj-model}
    \hat{\theta} = \min_{\theta} \sum_{i=1}^M \left(l(\hat{\bm{y}}_i, \bm{y}_i)\right) + 
    \lambda \|\bm{\theta}\|^2_F,
\end{align}
where $l(\cdot)$ denotes a classification loss function such as cross-entropy, $\hat{\bm{y}}_i = f(\bm{x}_i, \mathcal{N}(\bm{x}_i) | \hat{\theta}) \in \mathcal{R}^L$ denotes the model prediction for node $i$, and $\lambda$ is a hyper-parameter to balance the training loss and regularization term for preventing overfitting. It should be noted that $\hat{\bm{y}}_i$ is a probability distribution over the label space. The final classification $\hat{z}_i$ corresponds to the category with the largest probability, which is formulated as:
\begin{equation}\small
    \begin{aligned}
    \hat{z}_i = \arg\max_j~\hat{y}_{(i,j)}, j \leq L,
    \end{aligned}
\end{equation}
where $\hat{y}_{(i,j)}$ is the $j$-th entry of $\hat{\bm{y}}_i$. In the following, the mention of prediction and classification mean the predicted probability distribution ($\hat{\bm{y}}_i$) and category ($\hat{z}_i$), respectively. Besides, the subscript $i$ will be omitted for briefness.

\subsection{Cause-effect View} 
\paragraph{Causal graph.} Causal graph is a directed acyclic graph to describe a data generation process~\cite{pearl2009causality}, where nodes represent variables in the process, and edges represent the causal relations between variables. To facilitate analyzing the inference of GCN, \ie the generation process of the output, we abstract the inference of GCN as a causal graph (Figure~\ref{fig:causal_graph}(a)), which consists of four variables:
\begin{itemize}[leftmargin=*]
    \item $X$, which denotes the features of the target node. $\bm{x}$ is an instance of the variable.
    \item $N$, which denotes the neighbors of the target node, \eg $\mathcal{N}(\bm{x})$. The sample space of $N$ is the power set of all nodes in $G$.
    \item $\bar{X}$, which is the output of graph convolution at the last GCN layer.
    \item $\hat{Y}$, which denotes the GCN prediction, \ie the instance of $\hat{Y}$ is $\hat{\bm{y}}$.
\end{itemize}
Functionally speaking, the structure $X \longrightarrow \bar{X} \longleftarrow N$ represents the graph convolution where both the target node features and neighbor nodes directly affect the convolution output. The output of the graph convolution $\bar{X}$ then directly affects the model prediction, which is represented as $\bar{X} \longrightarrow \hat{Y}$
Note that there is a direct edge $X \longrightarrow \hat{Y}$, which means that $X$ directly affects the prediction. We include this direct edge for two considerations: 1) residual connection is widely used in GCN to prevent the over-smoothing issue~\cite{kipf2016semi}, which enables the features of the target node influence its prediction directly; 2) recent studies reveal the advantages of two-stage GCN where the model first makes prediction from each node's features; and then conducts graph convolution.

Recall that the conventional GCN inference, \ie the calculation of $\hat{\bm{y}}$, is typically a one-pass forward propagation of the GCN model with $\bm{x}$ and $\mathcal{N}(\bm{x})$ as inputs. 
Based on the causal graph, the procedure can be interpreted as Figure~\ref{fig:causal_graph}(b), where every variable obtains an instance (\eg $X = \bm{x}$). 
Apart from the new understanding of the conventional GCN inference, the causal theory provides analytical tools based on the causal graph, such as causal intervention~\cite{pearl2009causality}, which enable the in-depth analysis of the factors resulting in the prediction and further reasoning based on the prediction~\cite{pearl2019seven}. 

\paragraph{Causal intervention.} Our target is to assess whether the prediction on a target testing node faces the local structure discrepancy issue and further adjust the prediction to achieve adaptive locality. We resort to causal intervention to estimate the causal effect of target node's neighbors on the prediction (\ie the causal effect of $N=\mathcal{N}(\bm{x})$), which forcibly assigns an instance to a treatment variable. Formally, the causal effect $\bm{e} \in \mathbb{R}^L$ is defined as:
\begin{equation}\small
    \begin{aligned}
    \bm{e} &= f(\bm{x}, \mathcal{N}(\bm{x}) | \hat{\theta}) - f(\bm{x}, do(N = \emptyset) | \hat{\theta}),\notag \\
    &= f(\bm{x}, \mathcal{N}(\bm{x}) | \hat{\theta}) - f(\bm{x}, \emptyset | \hat{\theta}),\\ \notag
    &= \hat{\bm{y}} - \hat{\bm{y}}^s.
    \end{aligned}
\end{equation}
$do(N = \emptyset)$ represents a causal intervention which forcefully assigns a reference status of $N$, resulting in a post-intervention prediction $f(\bm{x}, do(N = \emptyset) | \hat{\theta})$ (see Figure~\ref{fig:causal_graph}(c)). Since $N$ does not have predecessor, $f(\bm{x}, do(N = \emptyset) | \hat{\theta}) = f(\bm{x}, \emptyset | \hat{\theta})$, which is denoted as $\hat{\bm{y}}^s \in \mathcal{R}^{L}$. Intuitively, the post-intervention prediction means: \textit{if the target node has no neighbor, what the prediction would be}. We believe that $\bm{e}$ provides clues for performing adaptive locality on the target node. For instance, we might adjust the original prediction, if the entries of $\bm{e}$ have abnormal large absolute values, which means that the local structure at the target node may not satisfy the homophily assumption~\cite{mcpherson2001birds}. Note that we take empty set as a representative reference status of $N = \mathcal{N}(\bm{x})$ since the widely usage of empty value as reference in causal intervention~\cite{pearl2009causality}, but can replace it with any subset of $\mathcal{N}(\bm{x})$ (see Section~\ref{ssec:factors}).



\subsection{Causal GCN Inference Mechanism}
The requirement of adaptive locality for the testing nodes pushes us to build up an additional mechanism, \ie CGI, to enhance the GCN inference stage. We have two main considerations for devising the mechanism: 1) 
the mechanism has to be learned from the data, instead of handcrafted, to enable its usage on different GCN models and different datasets. 2)
the mechanism should effectively capture the connections between the causal effect of $N=\mathcal{N}(\bm{x})$ and local structure discrepancy, \ie learning the patterns for adjusting the original prediction to improve the prediction accuracy.

\paragraph{$L$-way classification model.} 
A straight forward solution is devising the CGI as a $L$-way classification model that directly generates the final prediction according to the original prediction $\hat{\bm{y}}$, the post-intervention prediction $\hat{\bm{y}}^s$, and the causal effect $\bm{e}$. Formally,
\begin{equation}\small
    \begin{aligned}
    \bar{\bm{y}} = h(\hat{\bm{y}}, \hat{\bm{y}}^s, \bm{e} | \omega),
    \end{aligned}
\end{equation}
where $h(\cdot)$ denotes a $L$-way classifier parameterized by $\omega$ and $\bar{\bm{y}} \in \mathbb{R}^{L}$ denotes the final prediction. Similar to the training of GCN, we can learn the parameters of $h(\cdot)$ by optimizing classification loss over the labeled nodes, which is formulated as:
\begin{equation}\small
    \begin{aligned}\label{eq:obj-cgi-l}
    \hat{\bm{\omega}} = \min_{\bm{\omega}} \sum_{i=1}^M \left(l(\bar{\bm{y}}_i, \bm{y}_i)\right) + 
    \alpha \|\bm{\omega}\|^2_F,
    \end{aligned}
\end{equation}
where $\alpha$ is a hyperparameter to adjust the strength of regularization. 
Undoubtedly, this model can be easily developed and applied to any GCN. However, as optimized over the overall classification loss, the model will face the similar issue of attention mechanism~\cite{knyazev2019understanding}. To bridge this gap, it is essential to learn CGI under the awareness of whether a testing node encounters local structure discrepancy. 

\paragraph{Choice model.} 
Therefore, the inference mechanism should focus on the nodes with inconsistent classifications from $\hat{\bm{y}}$ and $\hat{\bm{y}}^s$, \ie $\hat{z} \neq \hat{z}^s$ where $\hat{z}$ is the original classification and $\hat{z}^s = \arg \max_j~\hat{y}_{(j)}^s, j \leq L$ is the post-intervention classification. That is, we let the CGI mechanism learn from nodes where accounting for neighbors causes the change of the classification.
To this end, we devise the inference mechanism as a \textit{choice model}, which is expected to make wise choice between $\hat{z}$ and $\hat{z}^s$ to eliminate the impact of local structure discrepancy. Formally,
\begin{equation}\small
    \begin{aligned}\label{eq:final-prediction}
    \bar{z} = \left\{
        \begin{aligned}
        & \hat{z}, \hat{p} \ge t,\\
        & \hat{z}^s, \hat{p} < t,
        \end{aligned}
    \right . 
    ~\hat{p} = g(\hat{\bm{y}}, \hat{\bm{y}}^s, \bm{e} | \bm{\eta}),
\end{aligned}
\end{equation}
where $g(\cdot)$ denotes a binary classifier with parameters of $\bm{\eta}$; the output of the classifier $p$ is used for making choice; and $t$ is the decision threshold, which depends on the classifier selected.

To learn the model parameters $\bm{\eta}$, we calculate the ground truth for making choice according to the correctness of $\hat{z}$ and $\hat{z}^s$. Formally, the choice training data of the binary classifier is:
\begin{equation}\small
    \begin{aligned}
    \mathcal{D} = \left\{(\bm{x}, p) | \hat{z} = z \cup \hat{z}^s = z \right\},~p = flag(\hat{z} = z),
\end{aligned}
\end{equation}
where $z$ denotes the correct category of node $\bm{x}$; $flag(\hat{z} = z) = 1$ if $\hat{z}$ equals to $z$, $flag(\hat{z} = z) = -1$ otherwise. The training of the choice model is thus formulated as:
\begin{equation}\small
    \begin{aligned}\label{eq:obj-cgi-b}
    \hat{\bm{\eta}} = \min_{\bm{\eta}} \sum_{(\bm{x}, p) \in \mathcal{D}} l(\hat{p}, p) + 
    \beta \|\bm{\eta}\|^2_F,
    \end{aligned}
\end{equation}
where $\beta$ is a hyperparameter to adjust the strength of regularization. 

\paragraph{Data sparsity.} Inevitably, the choice training data $\mathcal{D}$ will face sparsity issue for two reasons: 1) labeled nodes are limited in some applications; and 2) only a small portion of the labeled nodes satisfy the criteria of $\mathcal{D}$. To tackle the data sparsity issue, we have two main considerations: 1) the complexity of the binary classifier should be controlled strictly. Towards this end, we devise the choice model as a Support Vector Machine~\cite{pedregosa2011scikit} (SVM), since SVM only requires a few samples to serve as the support vectors to make choice. 2) the inputs of the binary classifier should free from the number of classes $L$, which can be large in some applications. To this end, we distill low dimensional and representative factors from the two predictions (\ie $\hat{\bm{y}}$ and $\hat{\bm{y}}^s$) and the causal effect $\bm{e}$) to serve as the inputs of the choice model, which is detailed in Section~\ref{ssec:factors}.

To summarize, as compared to conventional GCN inference, the proposed CGI has two main differences:
\begin{itemize}[leftmargin=*]
    \item In addition to the original prediction, CGI calls for causal intervention to further make a post-intervention prediction.
    \item CGI makes choice between the original prediction and post-intervention prediction with a choice model.
\end{itemize}

Below summarizes the slight change of GCN's training and inference schema to apply the proposed CGI:

\vspace{-0.3cm}
\begin{algorithm}[h]\small
	\caption{Applying CGI to GCN}  
	\label{algo:paradigm-cgi}
	\begin{algorithmic}[1]
		\Require Training data $\bm{X}$, $\bm{A}$, $\bm{Y}$. 
		
	    /* Training */
	    
	    \State Optimize Equation~(\ref{eq:obj-model}), obtaining GCN ($\hat{\bm{\theta}}$);  \algorithmiccomment{GCN training}
	    
	    \State Construct $\mathcal{D}$; \algorithmiccomment{Causal intervention}
	    
	    \State Optimize Equation~(\ref{eq:obj-cgi-b}), obtaining choice model (${\hat{\bm{\eta}}}$); \algorithmiccomment{CGI training}
	    
	    \State Return $\hat{\bm{\theta}}$ and ${\hat{\bm{\eta}}}$.

		/* Testing */
		\State Calculate $f(\bm{x}, \mathcal{N}(\bm{x}) | \hat{\bm{\theta}})$; \algorithmiccomment{Original prediction}
		
		\State Calculate $f(\bm{x}, \emptyset | \hat{\bm{\theta}})$; \algorithmiccomment{Post-intervention prediction}
		
		Calculate final classification with Equation (\ref{eq:final-prediction});
	\end{algorithmic}
\end{algorithm}
\setlength{\textfloatsep}{0.1cm}
\vspace{-0.4cm}

\subsection{Input Factors}~\label{ssec:factors}
To reduce the complexity of the choice model, we distill three types of factors as the input: \textit{causal uncertainty}, \textit{prediction confidence}, \textit{category transition}.

\paragraph{Causal uncertainty.} According to the homophily assumption~\cite{mcpherson2001birds}, the neighbors should not largely changes the prediction of the target node. Therefore, the target node may face the local structure discrepancy issue if the causal effect $\bm{e}$ of the neighbors has large variance. That is, the causal effect exhibits high uncertainty \wrt different reference values.
Inspired by the Monte Carlo uncertainty estimation, we resort to the variance of $\bm{e}$ to describe the causal uncertainty, which is formulated as:
\begin{align}\label{eq:mce}
    \bm{v} = var(\{f(\bm{x}, \mathcal{N}(\bm{x})_k | \hat{\theta}) | k \leq K \}),
\end{align}
where $\mathcal{N}(\bm{x})_k \subset \mathcal{N}(\bm{x})$, $var(\cdot)$ is an element-wise operation that calculates the variance on each class over the $K$ samples, and $\bm{v} \in \mathcal{R}^L$ denotes class-wise variance. In particular, we perform $K$ times of causal intervention with $N = \mathcal{N}(\bm{x})_k$ and then calculate the variance of the corresponding $K$ causal effects.
If an entry of $\bm{v}$ exhibits a large value, it reflects that minor changes on the subgraph structure can cause large changes on the prediction probability over the corresponds class. According to the original classification $\hat{z}$, we select the $\hat{z}$-th entry of $\bm{v}$ as a representative of the Monte Carlo causal effect uncertainty, which is termed as \textit{graph\_var}. 
In practice, we calculate the post-intervention predictions by repeating $K$ times of GCN inference with edge dropout~\cite{rong2019dropedge} applied, \ie each edge has a probability $\tau$ to be removed.



\paragraph{Prediction confidence.} There has been a surge of attention on using model predictions such as model distillation~\cite{hinton2015distilling} and self-supervised learning~\cite{chen2020big}. The intuition is that a larger probability indicates higher confidence on the classification. As such, a factor of prediction reliability is the prediction confidence, \ie trusting the prediction with higher confidence. Formally, we calculate two factors: \textit{self\_conf} ($\hat{\bm{y}}_{\hat{z}}$) and \textit{neighbor\_conf} ($\hat{\bm{y}}^s_{\hat{z}^s}$), respectively.

\paragraph{Category transition.} The distribution of edges over categories is not uniform \wrt: the ratio of \textit{intra-category connection} and \textit{inter-category connection}. Over the labeled nodes, we can calculate the distribution and form a category transition matrix $T$ where $T_{i,j}$ is the ratio of edges between category $i$ and $j$ to the edges connect category $i$. Figure~\ref{fig:cat_trans} illustrates 
an example on the OGB-arXiv dataset (raw normalized). We can see that the probability of intro-category connection (diagonal entries) varies in a large range ([0.19, 0.64]). The distribution of inter-category probability is also skewed. Intuitively, such probabilities can be clues for choosing the correct prediction, \eg $\hat{\bm{y}}$ might be trustworthy if $T_{\hat{z},\hat{z}}$ is high. To this end, we calculate four factors: \textit{self\_self} ($T_{\hat{z},\hat{z}}$), \textit{neighbor\_neighbor} ($T_{\hat{z}^s,\hat{z}^s}$), \textit{self\_neighbor} ($T_{\hat{z}, \hat{z}^s}$), and \textit{neighbor\_self} ($T_{\hat{z}^s,\hat{z}}$).

%% file: 5_exp.tex
\section{Experiments}
We conduct experiments on seven node classification datasets to answer the following research questions: 
\textbf{RQ1}: How effective is the proposed CGI model to resolve the local structure discrepancy issue?  \textbf{RQ2}: To what extend the proposed CGI facilitates node classification under different problem settings? \textbf{RQ3}: How do the distilled factors influence the effectiveness of the proposed CGI?

\begin{figure}[t]
	\centering
	\includegraphics[width=0.38\textwidth]{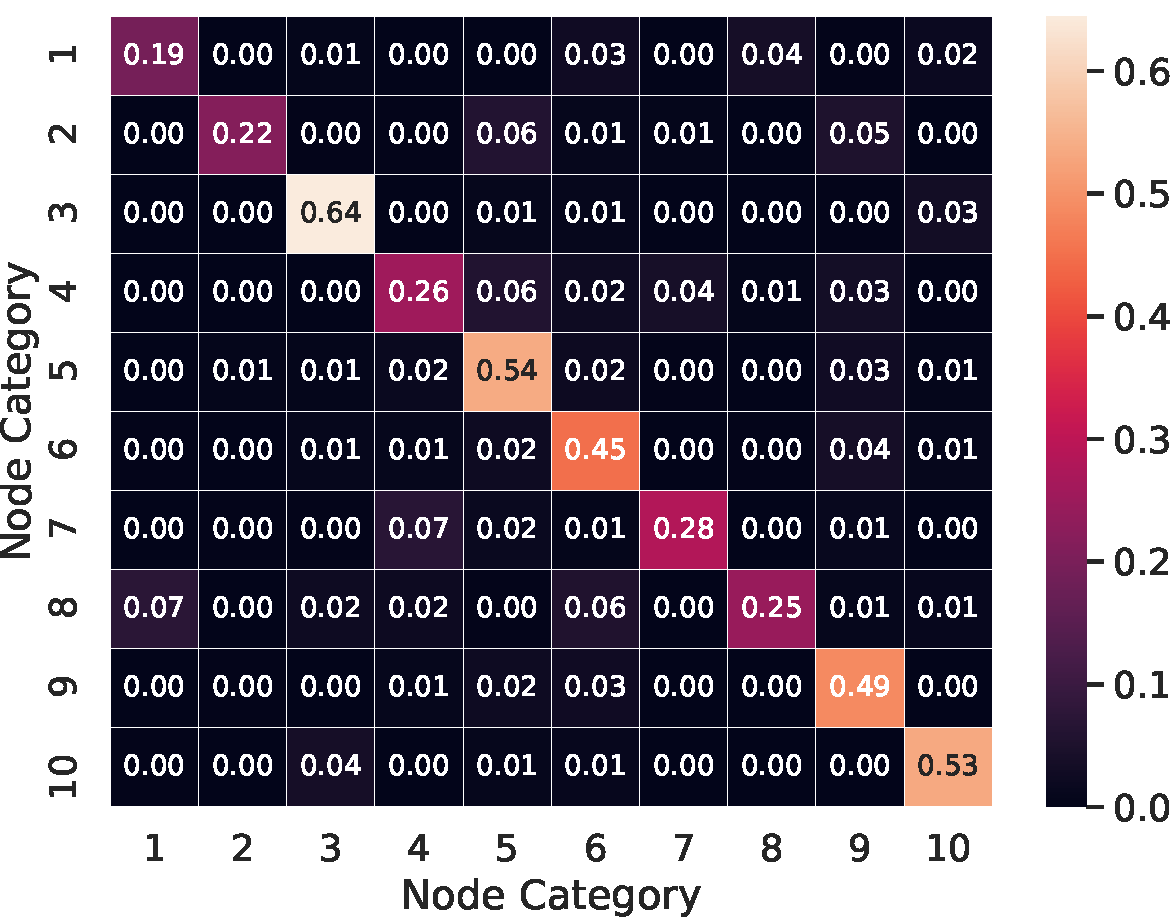}
	\vspace{-0.3cm}
	\caption{The category transition matrix on OGB-arXiv. To save space, we cut the number of categories to ten.
	}
	\label{fig:cat_trans}
\end{figure}

\subsection{Experimental Settings}
\subsubsection{Dataset}\label{sssec:dataset}
For the full-supervised settings, we use the widely used benchmark dataset of citation network, OGB-arXiv~\cite{OGB}, which represents papers and their citation relations as nodes and edges, respectively. Each node has 128 features generated by averaging the embeddings of words in its title and abstract, where the embeddings are learned by the skip-gram model~\cite{mikolov2013distributed}. Considering that the such old-fashioned text feature may not be representative, we replace the node features with a 768-dimensional vector extracted by feeding the title and abstract into RoBERTa~\cite{liu2019roberta} (12-layer
), where the representation of [CLS] token at the second last layer is selected. 

For the semi-supervised settings, we adopt three widely used citation networks, Cora, Citeseer, and Pubmed, and select the 20-shot data split released by~\cite{kipf2016semi}, where 500 and 1000 nodes are selected as validation and testing, 20 nodes from each category are labeled for training. Apart from the real-world graphs, we further created three synthetic ones based on Citeseer by intentionally adding cross-category edges on 50\% randomly selected nodes, which leads to local structure discrepancy between the poisoned nodes and the unaffected ones. Note that more cross-category edges lead to stronger discrepancy, making adaptive locality more critical for GCN models. In particular, according to the number of edges in the original Citeseer, we add 10\%, 30\%, and 50\% of cross-category edges, constructing Citeseer(10\%), Citeseer(30\%), and Citeseer(50\%). Note that the data split and node features are unchanged.

\subsubsection{Compared Methods} To justify the proposed CGI, we compare it with the representative GCN, including GraphSAGE~\cite{GraphSage}, GCN~\cite{kipf2016semi}, GAT~\cite{GAT}, JKNet~\cite{JKNet}, DAGNN~\cite{DAGNN}, and APPNP~\cite{APPNP}, which adopts normal inference. Apart from GCN models, we also test MLP, which discard the graph structure and treat node classification as normal text classification. Lastly, as CGI uses two predictions, we include an ensemble baseline which averages the prediction of APPNP and MLP.
For these models, we use the implementations on the OGB leaderboard\footnote{\url{https://ogb.stanford.edu/docs/leader_nodeprop/\#ogbn-arxiv}.}. If necessary, \eg under the transductive full-supervised setting, the hyper-parameters are tuned according to the settings in the original paper of the model. For the proposed CGI, we equipped the SVM with RBF kernel\footnote{\url{https://scikit-learn.org/stable/modules/svm.html}.} and apply CGI to APPNP. For the SVM, we tune two hyper-parameters $c$ and $\gamma$ through 5-fold cross-validation, \ie splitting the nodes in validation into 5 folds. In addition, for the MCE that estimates graph uncertainty, we set the number of repeats and the edge dropout ratio $\tau$ as 50 and 0.15.

\subsection{Effects of CGI (RQ1)} 
We first investigate to what extent the proposed CGI address the local structure discrepancy issue on the three synthetic datasets Citeseer(10\%), Citeseer(30\%), and Citeseer(50\%).
Table~\ref{tab:perf_synthetic} shows the performance of APPNP, APPNP\_Self, and APPNP\_CGI on the three datasets, where the final prediction is the original prediction (\ie $\hat{\bm{y}}$, the post-intervention prediction (\ie $\hat{\bm{y}}^s$), and the final prediction from CGI (\ie $\bar{\bm{y}}$), respectively. Note that the graph structure of the three datasets are different, the APPNP model trained on the datasets will thus be different. As such, the APPNP\_Self will get different performance on the three datasets, while the node features are unchanged. 
From the table, we have the following observations:
\begin{itemize}[leftmargin=*]
    \item In all cases, APPNP\_CGI outperforms APPNP, which validates the effectiveness of the proposed CGI. The performance gain is attributed to the further consideration of adaptive locality during GCN inference. In particular, the relative improvement over APPNP achieved by APPNP\_CGI ranges from 1.1\% to 7.2\% across the three datasets. The result shows that CGI achieves large improvement over compared conventional one-pass GCN inference as more cross-category edges are injected, \ie facing with more severe local structure discrepancy issue. The result further exhibits the capability of CGI to address the local structure discrepancy issue.
    
    \item As more cross-category edges being added, APPNP witnesses severe performance drop from the accuracy of 71.0\% to 64.2\%. This result is reasonable since GCN is vulnerable to cross-category edges which pushes the representations of node in different categories to be close~\cite{chen2019improving}. Considering that APPNP has considered adaptive locality during model training, this result validates that adjusting the GCN architecture is insufficient to address the local structure discrepancy issue.
    
    \item As to APPNP\_Self, the performance across the three datasets is comparable to each other. It indicates that the cross-category edges may not hinder the GCN to encode the association between target node features and the label. Therefore, the performance drop of APPNP when adding more cross-category edges is largely due to the improper neighbor aggregation without thorough consideration of the local structure discrepancy issue. 
    Furthermore, on Citeseer(50\%), the performance of APPNP\_Self is comparable to APPNP, which indicates that the effect of considering adaptive locality during training is limited if the discrepancy is very strong.
\end{itemize}

\subsection{Performance Comparison (RQ2)}
To further verify the proposed CGI, we conduct performance comparison under both full-supervised and semi-supervised settings on the real-world datasets.

\subsubsection{Semi-supervised setting.} 
We first investigate the effect of CGI under the semi-supervised setting by comparing the APPNP\_CGI with APPNP, APPNP\_Self, and APPNP\_Ensemble. The four methods corresponds to four inference mechanisms: 1) conventional one-pass GCN inference (APPNP); 2) causal intervention without consideration of graph structure (APPNP\_Self); 3) ensemble of APPNP and APPNP\_Self (APPNP\_Ensemble); and 4) the proposed CGI. Note that the four inference mechanisms are applied on the same APPNP model with exactly same model parameters. Table~\ref{tab:perf_semi-sup} shows the node classification performance on three real-world datasets: Cora, Citeseer, and Pubmed. From the table, we have the following observations: 
\begin{itemize}[leftmargin=*]
    \item On the three datasets, the performance of APPNP\_Self is largely worse than APPNP, \ie omitting graph structure during GCN inference witnesses sharp performance drop under semi-supervised setting, which shows the importance of considering neighbors. Note that the performance of APPNP\_Self largely surpasses the performance of MLP reported in \cite{kipf2016semi}, which highlights the difference between performing causal intervention $do(N = \emptyset)$ on a GCN model and the inference of MLP which is trained without the consideration of graph structure.
    \item In all cases, APPNP\_Ensemble performs worse than APPNP, which is one of the base models of the ensemble. The inferior performance of APPNP\_Ensemble is mainly because of the huge gap between the performance of APPNP and APPNP\_Self. From this results, we can conclude that, under semi-supervised setting, simply aggregating the original prediction and post-intervention prediction does not necessarily lead to better adaptive locality. Furthermore, the results validate the rationality of a carefully designed inference mechanism.
    \item In all cases, APPNP\_CGI achieves the best performance. The performance gain is attributed to the choice model, which further validates the effectiveness of the proposed CGI. That is, it is essential to an inference model from the data, which accounts for the causal analysis of the original prediction. 
    Moreover, this result reflects the potential of enhancing the inference mechanism of GCN for better decision making, especially the causality oriented analysis, which deserves further exploration in future research.
\end{itemize}

\begin{table}[t]
\centering
\resizebox{0.48\textwidth}{!}{%
\begin{tabular}{l|cccc}
\hline
Dataset & Citeseer(10\%) & Citeseer(30\%) & Citeseer(50\%)\\ \hline
APPNP & 71.0\% & 64.4\% & 64.2\% \\ 
APPNP\_Self & 65.1\% & 62.9\% & 64.3\% \\ 
APPNP\_CGI & \textbf{71.8\%} & \textbf{66.9\%} & \textbf{68.6\%} \\ \hline
RI & 1.1\% & 3.9\% & 7.2\% \\ \hline
\end{tabular}%
}
\caption{Performance of APPNP's original prediction, post-intervention prediction, and CGI prediction on the three synthetic datasets \wrt accuracy. RI means the relative improvement over APPNP achieved by APPNP\_CGI.}
\label{tab:perf_synthetic}
\vspace{-1.0cm}
\end{table}

\begin{table}[t]
\centering
\resizebox{0.38\textwidth}{!}{%
\begin{tabular}{l|ccc}
\hline
Dataset & Cora & Citeseer & Pubmed \\ \hline
APPNP & 81.8\% & 72.6\% & 79.8\% \\ 
APPNP\_Self & 69.3\% & 66.5\% & 75.9\% \\ 
APPNP\_Ensemble & 78.0\% & 71.4\% & 79.2\% \\ 
APPNP\_CGI & \textbf{82.3\%} & \textbf{73.7\%} & \textbf{81.0\%} \\ \hline
RI & 5.5\% & 2.8\% & 2.3\% \\ \hline
\end{tabular}%
}
\caption{Performance of APPNP with different inference mechanisms on three semi-supervised node classification datasets \wrt the classification accuracy. RI means the relative improvement of APPNP\_CGI over APPNP\_Ensemble.}
\label{tab:perf_semi-sup}
\vspace{-0.1cm}
\end{table}


\subsubsection{Full-supervised setting.} We then further investigate the effect of CGI under the full-supervised setting.
Note that we test the models under both inductive and transductive settings on the OGB-arXiv dataset. As OGB-arXiv is a widely used benchmark, we also test the baseline methods.  Table~\ref{tab:perf_full-sup} shows the node classification performance of the compared methods on the OGB-arXiv dataset \wrt accuracy. Apart from the RoBERTa features, we also report the performance of baseline models with the original Word2Vec features. From the table, we have the following observations:
\begin{itemize}[leftmargin=*]
    \item The performance gap between MLP and GCN models will be largely bridged when replacing the Word2Vec features with the more advanced RoBERTa features. In particular, the relative performance improvement of GCN models over MLP shrinks from 27.9\% to 3.7\%. The result raises a concern that the merit of GCN model might be unintentionally exaggerated~\cite{kipf2016semi} due to the low quality of node features. 
    \item Moreover, as compared to APPNP, DAGNN performs better as using the Word2Vec features, while performs worse when using the RoBERTa features. It suggests accounting for feature quality in future research that investigates the capability of GCN or compares different GCN models.
    \item As to RoBERTa features, APPNP\_Ensemble performs slightly better than its base models, \ie APPNP\_Self and APPNP. This result is different from the result in Table~\ref{tab:perf_semi-sup} under semi-supervised setting where the performance of APPNP\_Self is inferior. We thus believe that improving the accuracy of the post-intervention prediction will benefit GCN inference~\cite{zhou2012ensemble}. As averaging the prediction of APPNP\_Self and APPNP can also be seen as a choosing strategy by comparing model confidence, the performance gain indicates the benefit of considering adaptive locality during inference under full-supervised setting.
    \item APPNP\_CGI further outperforms APPNP\_Ensemble under both inductive and transductive settings, which is attributed to the choice model that learns to make choice from patterns of causal uncertainty, prediction confidence, and category transition factors. This result thus also shows the merit of characterizing the prediction of GCN models with the distilled factors. 
    \item In all cases, the model achieves comparable performance under the inductive setting and the transductive setting. We postulate that the local structure discrepancy between training and testing nodes in the OGB-arXiv dataset is weak, which is thus hard for CGI to achieve huge improvements.
    In the following, the experiment is focused on the inductive setting which is closer to real-world scenarios that aim to serve the upcoming nodes.
\end{itemize}

\subsection{In-depth Analysis (RQ3)}

\subsubsection{Effects of Distilled Factors}
\begin{figure}[t]
	\centering
	\includegraphics[width=0.48\textwidth]{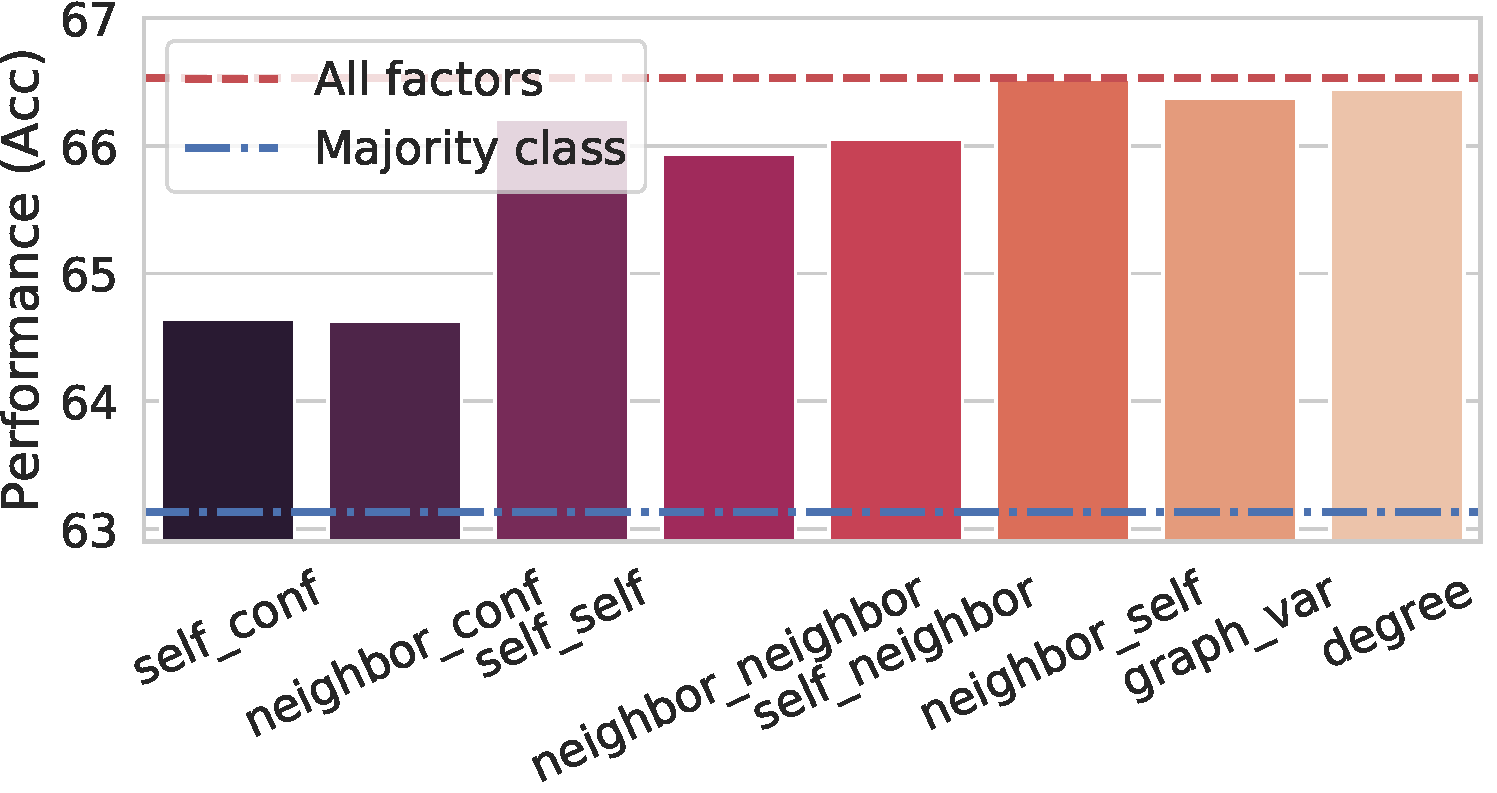}
	\vspace{-0.7cm}
	\caption{Illustration of factor influence on CGI.
	}
	\label{fig:factor_inf}
\end{figure}

We then study the effects of the distilled factors as the inputs of the choice model in CGI. In particular, we compare the factors \wrt the performance of CGI as removing one factor in each round, where lower performance indicates larger contribution of the factor. Note that we report the accuracy regarding whether CGI makes the correct choice for testing nodes, rather than the accuracy for node classification. That is to say, here we only consider ``conflict'' testing nodes where the two inferences of CGI (\ie APPNP and APPNP\_Self) have different classifications. Figure~\ref{fig:factor_inf} shows the performance on OGB-arXiv under the inductive setting, where 6,810 nodes among the 47,420 testing nodes are identified as the conflict nodes. We omit the results of other datasets under the semi-supervised setting for saving space, which have a close trend. 

From the figure, we have the following observations: 1) Discarding any factor will lead to performance drop as compared to the case with all factors as inputs of the choice model (\ie \textit{All factors}). This result indicates the effectiveness of the identified factors on characterizing GCN predictions which facilitate making the correct choice. 2) Among the factors, removing \textit{self\_conf} and \textit{neighbor\_conf} leads to the largest performance drop, showing that the confidence of prediction is the most informative factor regarding the reliability of the prediction. 3) In all cases, the performance of CGI surpasses the \textit{Majority class}, which always chooses the original GCN prediction, \ie CGI degrades to the conventional one-pass inference. The result further validates the rationality of additionally considering adaptive locality during GCN inference, \ie choosing between the original prediction and the post-intervention prediction without consideration of neighbors. Lastly, considering that the ``conflict'' nodes account for 14.4\% in the testing nodes (6,810/47,420) and the accuracy of CGI's choices is 66.53\%, there is still a large area for future exploration. 

\begin{table}[]
\centering
\resizebox{0.46\textwidth}{!}{%
\begin{tabular}{l|l|cc} \hline
Feature & Method & Inductive & Transductive \\ \hline
\multirow{7}{*}{\begin{tabular}[c]{@{}l@{}}Word2Vec\\ (128)\end{tabular}} & MLP & 55.84\% & 55.84\% \\
 & GraphSAGE & 71.43\%& 71.52\% \\
 & GCN & 71.83\% &  71.96\% \\
 & GAT &  71.93\% & 72.04\% \\
 & JKNet & \textbf{72.25\%} & \textbf{72.48\%} \\
 & DAGNN & \underline{72.07\%} & \underline{72.09\%} \\
 & APPNP & 71.61\% & 71.67\% \\ \hline
\multirow{6}{*}{\begin{tabular}[c]{@{}l@{}}RoBERTa\\ (768)\end{tabular}} & JKNet & 75.59\% & 75.54\% \\
 & MLP & 72.26\%  & 72.26\% \\
 & DAGNN &  74.93\% &  74.83\%\\
 & APPNP & 75.74\% & 75.61\% \\ \cline{2-4}
 & APPNP\_Self & 73.43\%  & 73.38\% \\
 & APPNP\_Ensemble & \underline{76.26\%} &  \underline{75.86\%}\\ 
 & APPNP\_CGI & \textbf{76.52\%} & \textbf{76.07\%} \\ \hline
\end{tabular}%
}
\caption{Performance comparison under full-supervised settings. We use bold font and underline to highlight the best and second best performance under each setting.}
\label{tab:perf_full-sup}
\vspace{-0.8cm}
\end{table}

\begin{figure*}[t]
	\centering
	\mbox{
		\subfigure[MCE]{
			\label{fig:group_stdev}
			\includegraphics[width=0.37\textwidth]{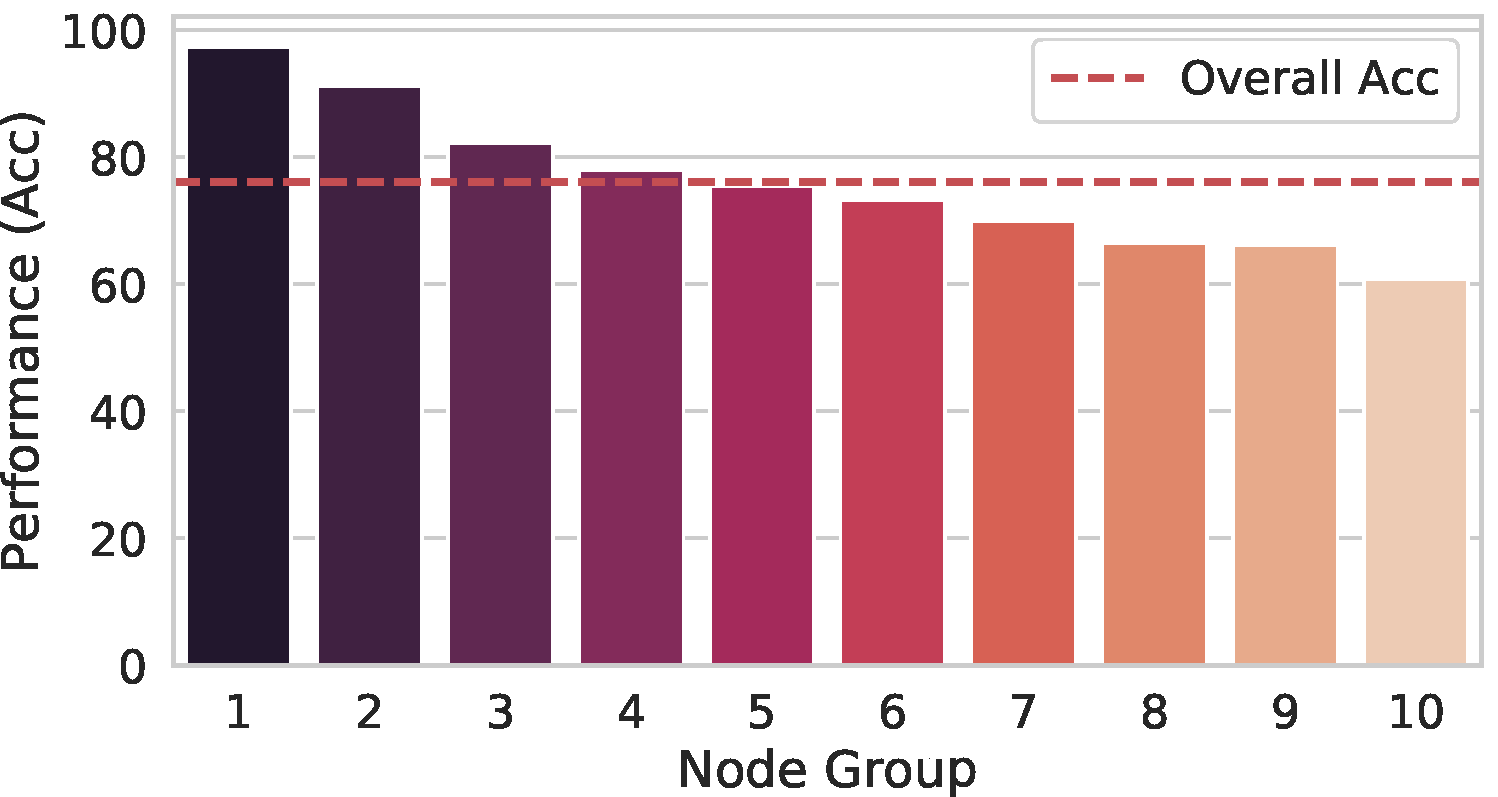}
		}
		\hspace{-0.1in}
		\subfigure[Confidence (Conf)]{
			\label{fig:group_conf}
			\includegraphics[width=0.37\textwidth]{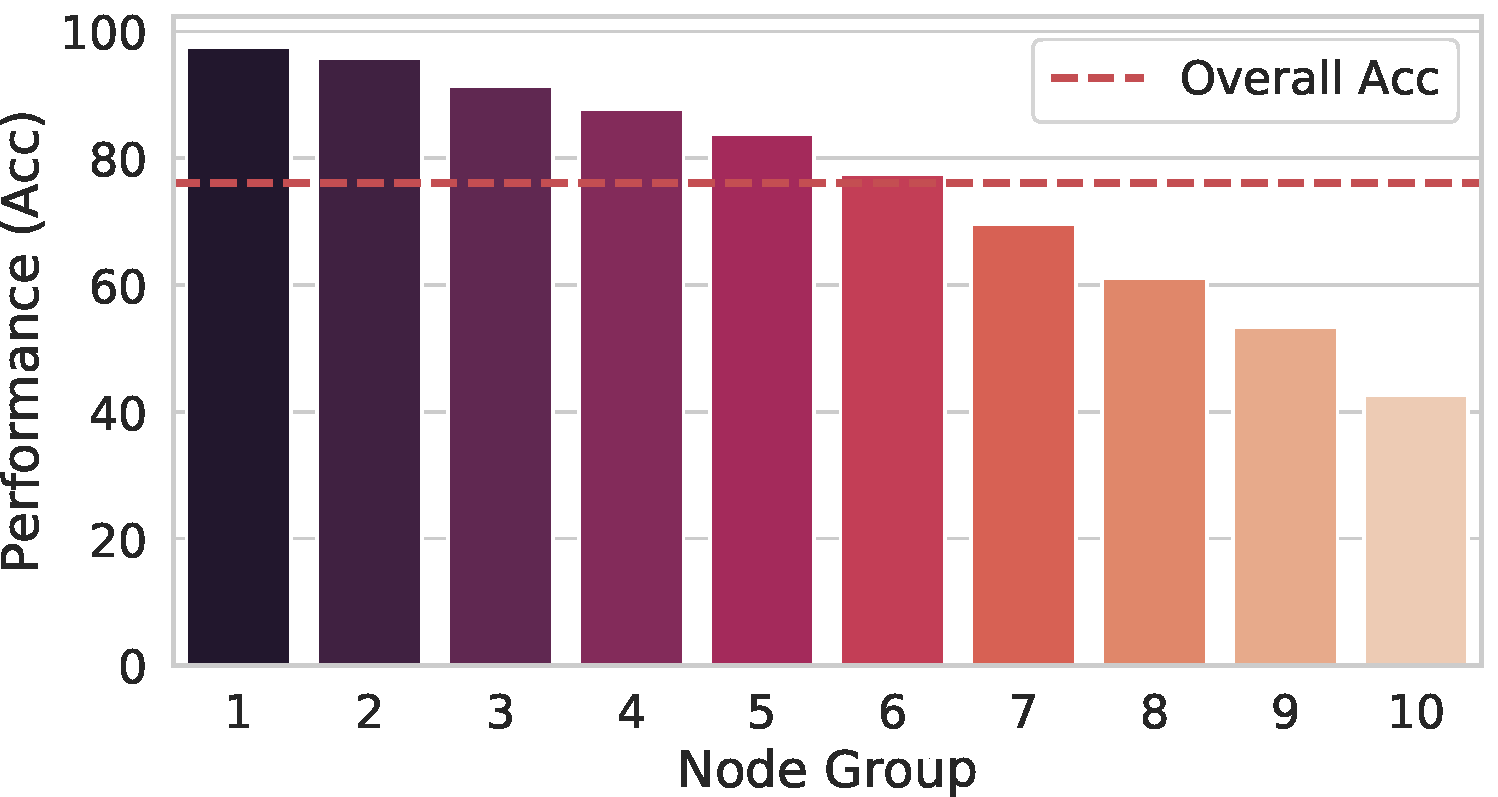}
		}
		\hspace{-0.1in}
		\subfigure[Conf and MCE redundancy]{
			\label{fig:group_ratio}
			\includegraphics[width=0.25\textwidth]{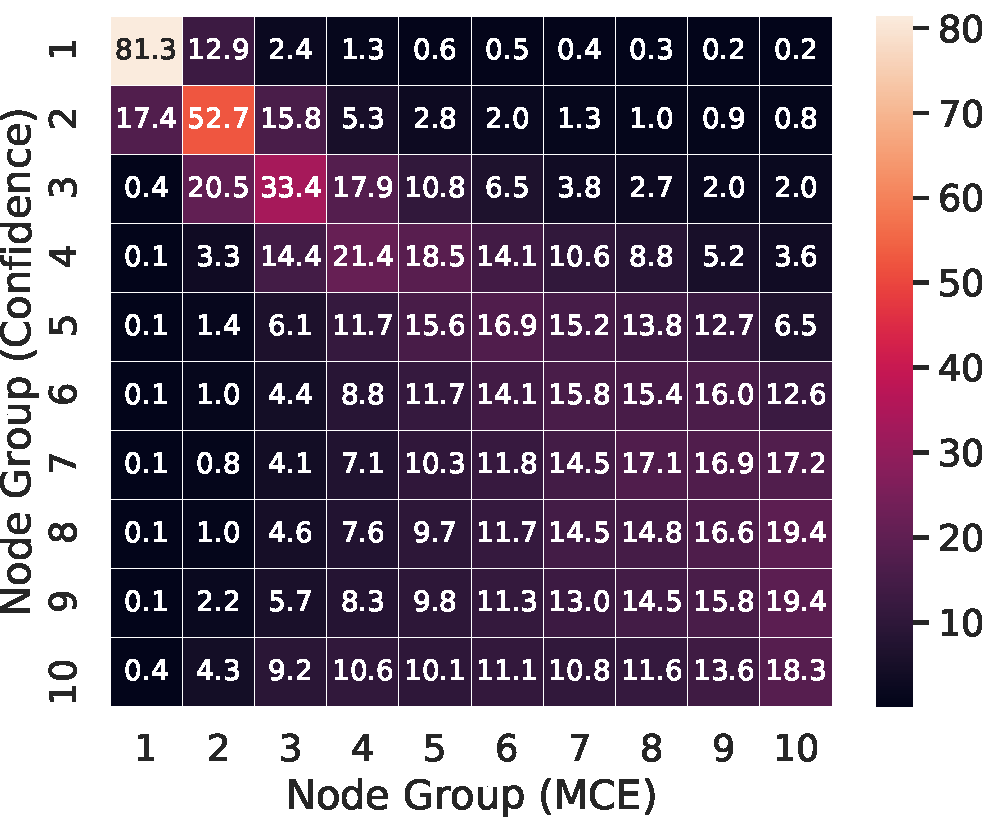}
		}
	}
	\vspace{-0.4cm}
	\caption{Group-wise illustration of the causal uncertainty and the confidence of APPNP prediction, where MCE stands for the Monte Carlo causal effect uncertainty estimation (\ie \textit{graph\_var}.)
	}
	\vspace{-0.2cm}
	\label{fig:graph_uncertainty}
\end{figure*}

\subsubsection{Study on causal uncertainty.} Recall that we propose a Monte Carlo causal effect uncertainty (MCE) estimation to estimate the uncertainty of neighbors' causal effect. We then investigate to what extent the MCE sheds light on the correctness of GCN prediction. Figure~\ref{fig:group_stdev} shows the group-wise performance of APPNP on OGB-arXiv where the testing nodes are ranked according to the value of graph\_var in an ascending order and split into ten groups with equal size. Note that we select OGB-arXiv for its relatively large scale where the testing set includes 47,420 nodes. From the figure, we can see a clear trend that the classification performance decreases as the MCE increases. It means that the calculated MCE is informative for the correctness of GCN predictions. For instance, a prediction has higher chance to be correct if its MCE is low. 

As a reference, in Figure~\ref{fig:group_conf}, we further depict the group-wise performance \wrt the prediction confidence (\ie neighbor\_conf). In particular, the testing nodes are ranked according to the value of neighbor\_conf in a descending order. As can be seen, there is also a clear trend of prediction performance regarding the confidence, \ie the probability of being correct is higher if APPNP is more confident on the prediction. To investigate whether MCE and GCN confidence are redundant, we further calculate the overlap ratio between the groups split by graph\_var and the ones split by neighbor\_conf. Figure~\ref{fig:group_ratio} illustrates the matrix of overlap ratios. As can be seen, the weights are not dedicated on the diagonal entries. In particular, there are only two group pairs with overlap ratio higher than 0.5, which means that the MCE reveals the property of GCN prediction complementary to the confidence. That is, causal analysis indeed characterizes GCN predictions from distinct perspectives.

%% file: 3_pilot.tex
\subsubsection{Training with trustworthiness signal}\label{sec:pilot}
\begin{table}[]
\centering
\resizebox{0.42\textwidth}{!}{%
\begin{tabular}{l|ccc}
\hline
Model & APPNP & JKNet & DAGNN \\ \hline
Self+Neighbor & 76.03 & 75.69 & 75.28 \\ 
Self+Neighbor\_Trust & 78.30 & 75.71 & 78.61 \\ 
Self+Neighbor\_Bound & 81.40 & 81.96 & 82.03 \\ \hline
\end{tabular}%
}
\caption{Node classification performance of three GCN models: APPNP, JKNet, and DAGNN on OGB-arXiv.}
\label{tab:pilot_study}
\end{table}

\begin{figure}[t]
	\centering
	\includegraphics[width=0.42\textwidth]{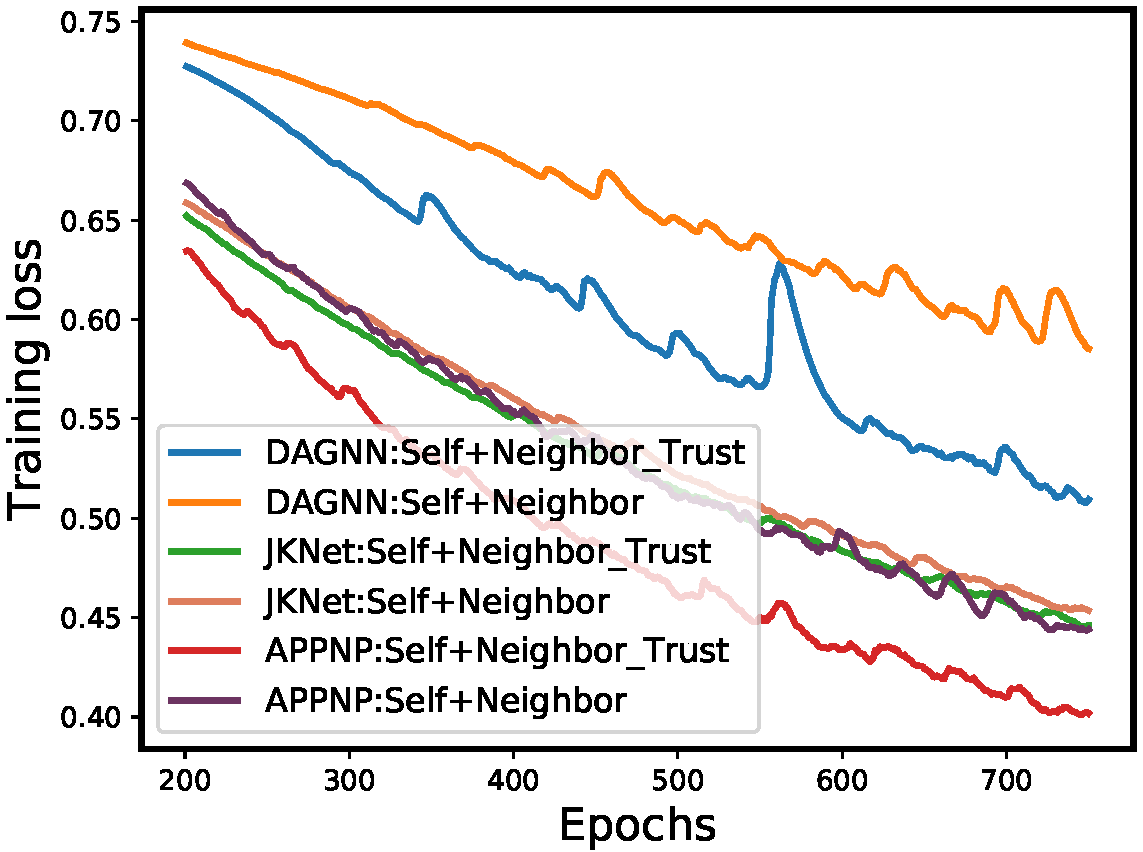}
	\vspace{-0.3cm}
	\caption{Training loss on OGB-arXiv under the Self+Neighbor and Self+Neighbor\_Trust configurations.
	}
	\label{fig:loss}
\end{figure}

To further investigate the benefit of performing adaptive locality during model inference, we further conduct a study on OGB-arXiv to test whether the GCN equipped with adaptive locality modules can really assess the trustworthiness of neighbors. Three representative GCN models with consideration of adaptive locality, APPNP, JKNet, and DAGNN, are tested under three different configurations:
\begin{itemize}[leftmargin= *]
    \item \textit{Self+Neighbor}: This is the standard configuration of GCN model that accounts for the graph structure, \ie trusting neighbors.
    \item \textit{Self+Neighbor\_Trust}: As compared to Self+Neighbor, a trustworthy feature is associated with each node, which indicates the ``ground truth'' of trusting self or neighbors. In particular, we train the GCN model, infer the original prediction and the post-intervention prediction, calculate the trustworthy feature according to Equation (10) (\ie $p$). For the nodes where the original classification and the post-intervention classification are equal, we set the value as 0. By explicitly incorporating such value as a node feature, it should be easy for GCN to learn for properly performing adaptive locality if it works properly.
    
    \item As a reference, we study the GCNs in an ideal case, named \textit{Self+Neighbor\_Bound}, where the trustworthy feature is also given when performing adaptive locality during model inference.
\end{itemize}

Table~\ref{tab:pilot_study} shows the model performance under the node classification setting of inductive full-supervised learning. 
From the table, we have the following observations:
\begin{itemize}[leftmargin=*]
    \item As compared to Self+Neighbor, all the three models, especially APPNP and DAGNN, achieve better performance under the configuration of Self+Neighbor\_Trust. It indicates a better usage of the graph structure, which is attributed to the trustworthy feature. The result thus highlights the importance of modeling neighbor trustworthiness and performing adaptive locality.
    
    \item However, there is a large gap between Self+Neighbor\_Trust and Self+Neighbor\_Bound, showing the underuse of the trustworthy feature by the current adaptive locality methods. We postulate the reason to be the gap between the training objective, \ie associating node representation with label, and the target of identifying trustworthy neighbors, which is the limitation of considering adaptive locality in model training. The performance under Self+Neighbor\_Bound also reveals the potential of considering adaptive locality in model inference.
\end{itemize}

Furthermore, we study the impact of trustworthy feature on model training. Figure~\ref{fig:loss} illustrates the training loss along the training procedure of the tested GCNs under the configuration of Self+Neighbor and Self+Neighbor\_Trust. It should be noted that we select the period from 200 to 750 epochs for better visualization. From the figure, we can see that, in all the three cases, the loss of GCN under Self+Neighbor\_Trust is smaller than that under Self+Neighbor. The result shows that the trustworthy feature facilitates the GCN model fitting the training data, \ie capturing the correlation between the node label and the node features as well as the graph structure. However, the adaptive locality module, especially graph attention, is distracted from the target of assessing neighbor trustworthiness. Theoretically, the graph attention can achieve the target by simply recognizing the value of the trustworthy feature from the inputs. For instance, the hop attention in DAGNN should highlight the target node representation at layer 0 if the trustworthy feature is 1. 


%% file: 6_rel.tex
\section{Related Work}
\paragraph{Graph Convolutional Network.}
According to the format of the convolution operations, existing GCN models can be divided into two categories: spatial GCN and spectral GCN~\cite{zhang2018deep}. Spectral GCN is defined as performing convolution operations in the Fourier domain with spectral node representations~\cite{bruna2013spectral,defferrard2016convolutional,kipf2016semi,liao2018lanczosnet,xu2018graph}. For instance, Bruna \etal \cite{bruna2013spectral} perform convolution over the eigenvectors of graph Laplacian which are treated as the Fourier basis. 
Due to the high computational cost of the eigen-decomposition, a line of spectral GCN research has been focused on accelerating the eigen-decomposition with different approximation techniques~\cite{defferrard2016convolutional,kipf2016semi,liao2018lanczosnet,xu2018graph}. 
However, applying such spectral GCN models on large graphs still raises unaffordable memory cost, which hinders the their practical research. 

To some extent, the attention on GCN research has been largely dedicated on the spatial GCN, which performs convolution operations directly over the graph structure by aggregating the features from spatially close neighbors to a target node~\cite{atwood2016diffusion,GraphSage,kipf2016semi,GAT,xinyi2018capsule,velickovic2018deep,xu2018powerful}. 
This line of research mainly focuses on the development of the neighbor aggregation operation. For instance, 
Kipf and Welling~\cite{kipf2016semi} propose to use a linear aggregator (\ie weighted sum) that uses the reverse of node degree as the coefficient. 
In addition to aggregating information from directly connected neighbors, augmented aggregators also account for multi-hop neighbors~\cite{GIL,jia2020redundancy}. 
Moreover, non-linear aggregators are also employed in spatial GCNs such as capsule~\cite{velickovic2018deep} and Long Short-Term Memory (LSTM)~\cite{GraphSage}.  
Besides, the general spatial GCN designed for simple graphs is extended to graphs with heterogeneous nodes~\cite{wang2019heterogeneous} and temporal structure~\cite{park2019exploiting}.
Beyond model design, there are also studies on the model capability analysis~\cite{xu2018powerful}, model explanation~\cite{ying2019gnnexplainer}, and training schema~\cite{DBLP:conf/iclr/HuLGZLPL20}. 

However, most of the existing studies focus on the training stage and blindly adopt the one-pass forward propagation for GCN inference. This work is in an orthogonal direction, which improve the inference performance with an causal inference mechanism so as to better solve the local structure discrepancy issue. Moreover, to the best of our knowledge, this work is the first to introduce the causal intervention and causal uncertainty into GCN inference.

\paragraph{Adaptive Locality.} Amongst the GCN research, a surge of attention has been especially dedicated to solving the over-smoothing issue~\cite{li2018deeper}. Adaptive locality has become the promising solution to alleviate the over-smoothing issue, which is typically achieved by the attention mechanism~\cite{GAT,JKNet,wang2019heterogeneous,yun2019graph,wang2020direct,wang2019improving,chen2019semisupervised,chen2020zero} or residual connection~\cite{kipf2016semi,chen2020simple,li2020deepergcn}. Along the line of research on attention design, integrating context information into the calculation of attention weight is one of the most popular techniques. For instance, Wang \etal~\cite{wang2020direct} treats the neighbors at different hops as augmentation of attention inputs. Moreover, to alleviate the issue of lacking direct supervision, Wang \etal~\cite{wang2019improving} introduce additional constraints to facilitate attention learning.
Similar as Convolutional Neural Networks, residual connection has also been introduced to original design of GCN~\cite{kipf2016semi}, which connects each layer to the output directly. In addition to the vanilla residual connection, the revised versions are also introduced such as the pre-activation residual~\cite{li2020deepergcn} and initial residual~\cite{chen2020simple}. Besides, the concept of inception module is also introduced to GCN model~\cite{kazi2019inceptiongcn}, which incorporates graph convolutions with different receptive fields. For the existing methods, the adaptive locality mechanism is fixed once the GCN model is trained. Instead, this work explores adaptive locality during model inference, which is in an orthogonal direction.

\paragraph{Causality-aware Model Prediction.} A surge of attention is being dedicated to incorporating causality into the ML schema~\cite{Yang2021DVMR,zhang2020devlbert,nan2021interventional,zhang2021cause,yue2020interventional,hu2021distilling}. A line of research focuses on enhancing the inference stage of ML model from the cause-effect view~\cite{yue2021counterfactual,wang2020click,niu2020counterfactual,tang2020long}. This work differs from them for two reasons: 1) none of the existing work studies GCN; and 2) we learn a choice model to make final prediction from causal intervention results instead of performing a heuristic causal inference for making final prediction.

%% file: 7_con.tex
\section{Conclusion}
This paper revealed that learning an additional model component such as graph attention is insufficient for addressing the local structure discrepancy issue of GCN models. Beyond model training, we explored the potential of empowering the GCNs with adaptive locality ability during the inference. In particular, we proposed an causal inference mechanism, which leverages the theory of causal intervention, generating a post intervention prediction when the model only trusts own features, and makes choice between the post intervention prediction and original prediction. A set of factors are identified to characterize the predictions and taken as inputs for choosing the final prediction. Especially, we proposed the Monte Carlo estimation of neighbors' causal effect. Under three common settings for node classification, we conducted extensive experiments on seven datasets, justifying the effectiveness of the proposed CGI.

In the future, we will test the proposed CGI on more GCN models such as GAT, JKNet, and DAGNN. Moreover, we will extend the proposed CGI from node classification to other graph analytic tasks, such as link prediction~\cite{ren2017social}. In addition, following the original Monte Carlo estimation, we would like to complete the mathematical derivation of Equation~\ref{eq:mce}, \ie how the graph\_var approximates the variance of the causal effect distribution.
Lastly, we will explore how the property of the choice model $g(\cdot)$ influences the effectiveness of CGI, \eg whether non-linearity is necessary.

%% file: 0_main.bbl

\begin{thebibliography}{66}


\ifx \showCODEN    \undefined \def \showCODEN     #1{\unskip}     \fi
\ifx \showDOI      \undefined \def \showDOI       #1{#1}\fi
\ifx \showISBNx    \undefined \def \showISBNx     #1{\unskip}     \fi
\ifx \showISBNxiii \undefined \def \showISBNxiii  #1{\unskip}     \fi
\ifx \showISSN     \undefined \def \showISSN      #1{\unskip}     \fi
\ifx \showLCCN     \undefined \def \showLCCN      #1{\unskip}     \fi
\ifx \shownote     \undefined \def \shownote      #1{#1}          \fi
\ifx \showarticletitle \undefined \def \showarticletitle #1{#1}   \fi
\ifx \showURL      \undefined \def \showURL       {\relax}        \fi
\providecommand\bibfield[2]{#2}
\providecommand\bibinfo[2]{#2}
\providecommand\natexlab[1]{#1}
\providecommand\showeprint[2][]{arXiv:#2}

\bibitem[\protect\citeauthoryear{Atwood and Towsley}{Atwood and
  Towsley}{2016}]%
        {atwood2016diffusion}
\bibfield{author}{\bibinfo{person}{James Atwood} {and} \bibinfo{person}{Don
  Towsley}.} \bibinfo{year}{2016}\natexlab{}.
\newblock \showarticletitle{Diffusion-convolutional neural networks}. In
  \bibinfo{booktitle}{\emph{NeurIPS}}. \bibinfo{pages}{1993--2001}.
\newblock


\bibitem[\protect\citeauthoryear{Bruna, Zaremba, Szlam, and LeCun}{Bruna
  et~al\mbox{.}}{2014}]%
        {bruna2013spectral}
\bibfield{author}{\bibinfo{person}{Joan Bruna}, \bibinfo{person}{Wojciech
  Zaremba}, \bibinfo{person}{Arthur Szlam}, {and} \bibinfo{person}{Yann
  LeCun}.} \bibinfo{year}{2014}\natexlab{}.
\newblock \showarticletitle{Spectral networks and locally connected networks on
  graphs}.
\newblock \bibinfo{journal}{\emph{ICLR}} (\bibinfo{year}{2014}).
\newblock


\bibitem[\protect\citeauthoryear{Chen, Lin, Li, Li, Zhou, and Sun}{Chen
  et~al\mbox{.}}{2020c}]%
        {chen2019measuring}
\bibfield{author}{\bibinfo{person}{Deli Chen}, \bibinfo{person}{Yankai Lin},
  \bibinfo{person}{Wei Li}, \bibinfo{person}{Peng Li}, \bibinfo{person}{Jie
  Zhou}, {and} \bibinfo{person}{Xu Sun}.} \bibinfo{year}{2020}\natexlab{c}.
\newblock \showarticletitle{Measuring and Relieving the Over-smoothing Problem
  for Graph Neural Networks from the Topological View}.
\newblock \bibinfo{journal}{\emph{AAAI}} (\bibinfo{year}{2020}),
  \bibinfo{pages}{3438--3445}.
\newblock


\bibitem[\protect\citeauthoryear{Chen, Liu, Lin, Li, Zhou, Su, and Sun}{Chen
  et~al\mbox{.}}{2019b}]%
        {chen2019improving}
\bibfield{author}{\bibinfo{person}{Deli Chen}, \bibinfo{person}{Xiaoqian Liu},
  \bibinfo{person}{Yankai Lin}, \bibinfo{person}{Peng Li}, \bibinfo{person}{Jie
  Zhou}, \bibinfo{person}{Qi Su}, {and} \bibinfo{person}{Xu Sun}.}
  \bibinfo{year}{2019}\natexlab{b}.
\newblock \showarticletitle{Improving Node Classification by Co-training Node
  Pair Classification: A Novel Training Framework for General Graph Neural
  Networks}.
\newblock \bibinfo{journal}{\emph{arXiv preprint arXiv:1911.03904}}
  (\bibinfo{year}{2019}).
\newblock


\bibitem[\protect\citeauthoryear{Chen, Dong, Wang, Feng, Wang, and He}{Chen
  et~al\mbox{.}}{2020a}]%
        {chen2020bias}
\bibfield{author}{\bibinfo{person}{Jiawei Chen}, \bibinfo{person}{Hande Dong},
  \bibinfo{person}{Xiang Wang}, \bibinfo{person}{Fuli Feng},
  \bibinfo{person}{Meng Wang}, {and} \bibinfo{person}{Xiangnan He}.}
  \bibinfo{year}{2020}\natexlab{a}.
\newblock \showarticletitle{Bias and Debias in Recommender System: A Survey and
  Future Directions}.
\newblock \bibinfo{journal}{\emph{arXiv preprint arXiv:2010.03240}}
  (\bibinfo{year}{2020}).
\newblock


\bibitem[\protect\citeauthoryear{Chen, Pan, Wei, Wang, Ngo, and Chua}{Chen
  et~al\mbox{.}}{2020d}]%
        {chen2020zero}
\bibfield{author}{\bibinfo{person}{Jingjing Chen}, \bibinfo{person}{Liangming
  Pan}, \bibinfo{person}{Zhipeng Wei}, \bibinfo{person}{Xiang Wang},
  \bibinfo{person}{Chong-Wah Ngo}, {and} \bibinfo{person}{Tat-Seng Chua}.}
  \bibinfo{year}{2020}\natexlab{d}.
\newblock \showarticletitle{Zero-shot ingredient recognition by
  multi-relational graph convolutional network}. In
  \bibinfo{booktitle}{\emph{Proceedings of the AAAI Conference on Artificial
  Intelligence}}, Vol.~\bibinfo{volume}{34}. \bibinfo{pages}{10542--10550}.
\newblock


\bibitem[\protect\citeauthoryear{Chen, Wei, Huang, Ding, and Li}{Chen
  et~al\mbox{.}}{2020e}]%
        {chen2020simple}
\bibfield{author}{\bibinfo{person}{Ming Chen}, \bibinfo{person}{Zhewei Wei},
  \bibinfo{person}{Zengfeng Huang}, \bibinfo{person}{Bolin Ding}, {and}
  \bibinfo{person}{Yaliang Li}.} \bibinfo{year}{2020}\natexlab{e}.
\newblock \showarticletitle{Simple and deep graph convolutional networks}.
\newblock \bibinfo{journal}{\emph{ICML}} (\bibinfo{year}{2020}).
\newblock


\bibitem[\protect\citeauthoryear{Chen, Kornblith, Swersky, Norouzi, and
  Hinton}{Chen et~al\mbox{.}}{2020b}]%
        {chen2020big}
\bibfield{author}{\bibinfo{person}{Ting Chen}, \bibinfo{person}{Simon
  Kornblith}, \bibinfo{person}{Kevin Swersky}, \bibinfo{person}{Mohammad
  Norouzi}, {and} \bibinfo{person}{Geoffrey Hinton}.}
  \bibinfo{year}{2020}\natexlab{b}.
\newblock \showarticletitle{Big self-supervised models are strong
  semi-supervised learners}.
\newblock \bibinfo{journal}{\emph{arXiv preprint arXiv:2006.10029}}
  (\bibinfo{year}{2020}).
\newblock


\bibitem[\protect\citeauthoryear{Chen, Gu, Ren, He, Xie, Guo, Yin, and
  Zhang}{Chen et~al\mbox{.}}{2019a}]%
        {chen2019semisupervised}
\bibfield{author}{\bibinfo{person}{Weijian Chen}, \bibinfo{person}{Yulong Gu},
  \bibinfo{person}{Zhaochun Ren}, \bibinfo{person}{Xiangnan He},
  \bibinfo{person}{Hongtao Xie}, \bibinfo{person}{Tong Guo},
  \bibinfo{person}{Dawei Yin}, {and} \bibinfo{person}{Yongdong Zhang}.}
  \bibinfo{year}{2019}\natexlab{a}.
\newblock \showarticletitle{Semi-supervised User Profiling with Heterogeneous
  Graph Attention Networks}. In \bibinfo{booktitle}{\emph{IJCAI}}.
  \bibinfo{pages}{2116--2122}.
\newblock


\bibitem[\protect\citeauthoryear{Defferrard, Bresson, and
  Vandergheynst}{Defferrard et~al\mbox{.}}{2016}]%
        {defferrard2016convolutional}
\bibfield{author}{\bibinfo{person}{Micha{\"e}l Defferrard},
  \bibinfo{person}{Xavier Bresson}, {and} \bibinfo{person}{Pierre
  Vandergheynst}.} \bibinfo{year}{2016}\natexlab{}.
\newblock \showarticletitle{Convolutional neural networks on graphs with fast
  localized spectral filtering}. In \bibinfo{booktitle}{\emph{NeurIPS}}.
  \bibinfo{pages}{3844--3852}.
\newblock


\bibitem[\protect\citeauthoryear{Hamilton, Ying, and Leskovec}{Hamilton
  et~al\mbox{.}}{2017}]%
        {GraphSage}
\bibfield{author}{\bibinfo{person}{Will Hamilton}, \bibinfo{person}{Zhitao
  Ying}, {and} \bibinfo{person}{Jure Leskovec}.}
  \bibinfo{year}{2017}\natexlab{}.
\newblock \showarticletitle{Inductive representation learning on large graphs}.
  In \bibinfo{booktitle}{\emph{NeurIPS}}. \bibinfo{pages}{1024--1034}.
\newblock


\bibitem[\protect\citeauthoryear{He, Deng, Wang, Li, Zhang, and Wang}{He
  et~al\mbox{.}}{2020}]%
        {LightGCN}
\bibfield{author}{\bibinfo{person}{Xiangnan He}, \bibinfo{person}{Kuan Deng},
  \bibinfo{person}{Xiang Wang}, \bibinfo{person}{Yan Li},
  \bibinfo{person}{Yongdong Zhang}, {and} \bibinfo{person}{Meng Wang}.}
  \bibinfo{year}{2020}\natexlab{}.
\newblock \showarticletitle{LightGCN: Simplifying and Powering Graph
  Convolution Network for Recommendation}. In
  \bibinfo{booktitle}{\emph{SIGIR}}.
\newblock


\bibitem[\protect\citeauthoryear{Hinton, Vinyals, and Dean}{Hinton
  et~al\mbox{.}}{2015}]%
        {hinton2015distilling}
\bibfield{author}{\bibinfo{person}{Geoffrey Hinton}, \bibinfo{person}{Oriol
  Vinyals}, {and} \bibinfo{person}{Jeff Dean}.}
  \bibinfo{year}{2015}\natexlab{}.
\newblock \showarticletitle{Distilling the knowledge in a neural network}.
\newblock \bibinfo{journal}{\emph{arXiv preprint arXiv:1503.02531}}
  (\bibinfo{year}{2015}).
\newblock


\bibitem[\protect\citeauthoryear{Hu, Shi, Hao, and Bai}{Hu
  et~al\mbox{.}}{2020c}]%
        {hu2020residual}
\bibfield{author}{\bibinfo{person}{Guangyi Hu}, \bibinfo{person}{Chongyang
  Shi}, \bibinfo{person}{Shufeng Hao}, {and} \bibinfo{person}{Yu Bai}.}
  \bibinfo{year}{2020}\natexlab{c}.
\newblock \showarticletitle{Residual-Duet Network with Tree Dependency
  Representation for Chinese Question-Answering Sentiment Analysis}. In
  \bibinfo{booktitle}{\emph{Proceedings of the 43rd International ACM SIGIR
  Conference on Research and Development in Information Retrieval}}.
  \bibinfo{pages}{1725--1728}.
\newblock


\bibitem[\protect\citeauthoryear{Hu, Fey, Zitnik, Dong, Ren, Liu, Catasta, and
  Leskovec}{Hu et~al\mbox{.}}{2020a}]%
        {OGB}
\bibfield{author}{\bibinfo{person}{Weihua Hu}, \bibinfo{person}{Matthias Fey},
  \bibinfo{person}{Marinka Zitnik}, \bibinfo{person}{Yuxiao Dong},
  \bibinfo{person}{Hongyu Ren}, \bibinfo{person}{Bowen Liu},
  \bibinfo{person}{Michele Catasta}, {and} \bibinfo{person}{Jure Leskovec}.}
  \bibinfo{year}{2020}\natexlab{a}.
\newblock \showarticletitle{Open graph benchmark: Datasets for machine learning
  on graphs}.
\newblock \bibinfo{journal}{\emph{NeurIPS}} (\bibinfo{year}{2020}).
\newblock


\bibitem[\protect\citeauthoryear{Hu, Liu, Gomes, Zitnik, Liang, Pande, and
  Leskovec}{Hu et~al\mbox{.}}{2020b}]%
        {DBLP:conf/iclr/HuLGZLPL20}
\bibfield{author}{\bibinfo{person}{Weihua Hu}, \bibinfo{person}{Bowen Liu},
  \bibinfo{person}{Joseph Gomes}, \bibinfo{person}{Marinka Zitnik},
  \bibinfo{person}{Percy Liang}, \bibinfo{person}{Vijay~S. Pande}, {and}
  \bibinfo{person}{Jure Leskovec}.} \bibinfo{year}{2020}\natexlab{b}.
\newblock \showarticletitle{Strategies for Pre-training Graph Neural Networks}.
  In \bibinfo{booktitle}{\emph{{ICLR}}}.
\newblock


\bibitem[\protect\citeauthoryear{Hu, Tang, Miao, Hua, and Zhang}{Hu
  et~al\mbox{.}}{2021}]%
        {hu2021distilling}
\bibfield{author}{\bibinfo{person}{Xinting Hu}, \bibinfo{person}{Kaihua Tang},
  \bibinfo{person}{Chunyan Miao}, \bibinfo{person}{Xian-Sheng Hua}, {and}
  \bibinfo{person}{Hanwang Zhang}.} \bibinfo{year}{2021}\natexlab{}.
\newblock \showarticletitle{Distilling Causal Effect of Data in
  Class-Incremental Learning}.
\newblock \bibinfo{journal}{\emph{IEEE Conference on Computer Vision and
  Pattern Recognition}} (\bibinfo{year}{2021}).
\newblock


\bibitem[\protect\citeauthoryear{Jia, Lin, Ying, You, Leskovec, and Aiken}{Jia
  et~al\mbox{.}}{2020}]%
        {jia2020redundancy}
\bibfield{author}{\bibinfo{person}{Zhihao Jia}, \bibinfo{person}{Sina Lin},
  \bibinfo{person}{Rex Ying}, \bibinfo{person}{Jiaxuan You},
  \bibinfo{person}{Jure Leskovec}, {and} \bibinfo{person}{Alex Aiken}.}
  \bibinfo{year}{2020}\natexlab{}.
\newblock \showarticletitle{Redundancy-Free Computation for Graph Neural
  Networks}. In \bibinfo{booktitle}{\emph{SIGKDD}}. \bibinfo{pages}{997--1005}.
\newblock


\bibitem[\protect\citeauthoryear{Kazi, Shekarforoush, Krishna, Burwinkel,
  Vivar, Kort{\"u}m, Ahmadi, Albarqouni, and Navab}{Kazi et~al\mbox{.}}{2019}]%
        {kazi2019inceptiongcn}
\bibfield{author}{\bibinfo{person}{Anees Kazi}, \bibinfo{person}{Shayan
  Shekarforoush}, \bibinfo{person}{S~Arvind Krishna}, \bibinfo{person}{Hendrik
  Burwinkel}, \bibinfo{person}{Gerome Vivar}, \bibinfo{person}{Karsten
  Kort{\"u}m}, \bibinfo{person}{Seyed-Ahmad Ahmadi}, \bibinfo{person}{Shadi
  Albarqouni}, {and} \bibinfo{person}{Nassir Navab}.}
  \bibinfo{year}{2019}\natexlab{}.
\newblock \showarticletitle{InceptionGCN: receptive field aware graph
  convolutional network for disease prediction}. In
  \bibinfo{booktitle}{\emph{International Conference on Information Processing
  in Medical Imaging}}. Springer, \bibinfo{pages}{73--85}.
\newblock


\bibitem[\protect\citeauthoryear{Kipf and Welling}{Kipf and Welling}{2017}]%
        {kipf2016semi}
\bibfield{author}{\bibinfo{person}{Thomas~N Kipf} {and} \bibinfo{person}{Max
  Welling}.} \bibinfo{year}{2017}\natexlab{}.
\newblock \showarticletitle{Semi-supervised classification with graph
  convolutional networks}.
\newblock \bibinfo{journal}{\emph{ICLR}} (\bibinfo{year}{2017}).
\newblock


\bibitem[\protect\citeauthoryear{Klicpera, Bojchevski, and
  G{\"u}nnemann}{Klicpera et~al\mbox{.}}{2019}]%
        {APPNP}
\bibfield{author}{\bibinfo{person}{Johannes Klicpera},
  \bibinfo{person}{Aleksandar Bojchevski}, {and} \bibinfo{person}{Stephan
  G{\"u}nnemann}.} \bibinfo{year}{2019}\natexlab{}.
\newblock \showarticletitle{Predict then propagate: Graph neural networks meet
  personalized pagerank}.
\newblock \bibinfo{journal}{\emph{ICLR}} (\bibinfo{year}{2019}).
\newblock


\bibitem[\protect\citeauthoryear{Knyazev, Taylor, and Amer}{Knyazev
  et~al\mbox{.}}{2019}]%
        {knyazev2019understanding}
\bibfield{author}{\bibinfo{person}{Boris Knyazev}, \bibinfo{person}{Graham~W
  Taylor}, {and} \bibinfo{person}{Mohamed Amer}.}
  \bibinfo{year}{2019}\natexlab{}.
\newblock \showarticletitle{Understanding Attention and Generalization in Graph
  Neural Networks}. In \bibinfo{booktitle}{\emph{NeurIPS}}.
  \bibinfo{pages}{4204--4214}.
\newblock


\bibitem[\protect\citeauthoryear{Li, Xiong, Thabet, and Ghanem}{Li
  et~al\mbox{.}}{2020}]%
        {li2020deepergcn}
\bibfield{author}{\bibinfo{person}{Guohao Li}, \bibinfo{person}{Chenxin Xiong},
  \bibinfo{person}{Ali Thabet}, {and} \bibinfo{person}{Bernard Ghanem}.}
  \bibinfo{year}{2020}\natexlab{}.
\newblock \showarticletitle{Deepergcn: All you need to train deeper gcns}.
\newblock \bibinfo{journal}{\emph{arXiv preprint arXiv:2006.07739}}
  (\bibinfo{year}{2020}).
\newblock


\bibitem[\protect\citeauthoryear{Li, Han, and Wu}{Li et~al\mbox{.}}{2018}]%
        {li2018deeper}
\bibfield{author}{\bibinfo{person}{Qimai Li}, \bibinfo{person}{Zhichao Han},
  {and} \bibinfo{person}{Xiao-Ming Wu}.} \bibinfo{year}{2018}\natexlab{}.
\newblock \showarticletitle{Deeper insights into graph convolutional networks
  for semi-supervised learning}. In \bibinfo{booktitle}{\emph{AAAI}}.
\newblock


\bibitem[\protect\citeauthoryear{Liao, Zhao, Urtasun, and Zemel}{Liao
  et~al\mbox{.}}{2019}]%
        {liao2018lanczosnet}
\bibfield{author}{\bibinfo{person}{Renjie Liao}, \bibinfo{person}{Zhizhen
  Zhao}, \bibinfo{person}{Raquel Urtasun}, {and} \bibinfo{person}{Richard
  Zemel}.} \bibinfo{year}{2019}\natexlab{}.
\newblock \showarticletitle{LanczosNet: Multi-Scale Deep Graph Convolutional
  Networks}. In \bibinfo{booktitle}{\emph{ICLR}}.
\newblock


\bibitem[\protect\citeauthoryear{Linmei, Yang, Shi, Ji, and Li}{Linmei
  et~al\mbox{.}}{2019}]%
        {linmei2019heterogeneous}
\bibfield{author}{\bibinfo{person}{Hu Linmei}, \bibinfo{person}{Tianchi Yang},
  \bibinfo{person}{Chuan Shi}, \bibinfo{person}{Houye Ji}, {and}
  \bibinfo{person}{Xiaoli Li}.} \bibinfo{year}{2019}\natexlab{}.
\newblock \showarticletitle{Heterogeneous graph attention networks for
  semi-supervised short text classification}. In
  \bibinfo{booktitle}{\emph{EMNLP-IJCNLP}}. \bibinfo{pages}{4823--4832}.
\newblock


\bibitem[\protect\citeauthoryear{Liu, Gao, and Ji}{Liu et~al\mbox{.}}{2020}]%
        {DAGNN}
\bibfield{author}{\bibinfo{person}{Meng Liu}, \bibinfo{person}{Hongyang Gao},
  {and} \bibinfo{person}{Shuiwang Ji}.} \bibinfo{year}{2020}\natexlab{}.
\newblock \showarticletitle{Towards Deeper Graph Neural Networks}. In
  \bibinfo{booktitle}{\emph{SIGKDD}}. \bibinfo{pages}{338--348}.
\newblock


\bibitem[\protect\citeauthoryear{Liu, Ott, Goyal, Du, Joshi, Chen, Levy, Lewis,
  Zettlemoyer, and Stoyanov}{Liu et~al\mbox{.}}{2019}]%
        {liu2019roberta}
\bibfield{author}{\bibinfo{person}{Yinhan Liu}, \bibinfo{person}{Myle Ott},
  \bibinfo{person}{Naman Goyal}, \bibinfo{person}{Jingfei Du},
  \bibinfo{person}{Mandar Joshi}, \bibinfo{person}{Danqi Chen},
  \bibinfo{person}{Omer Levy}, \bibinfo{person}{Mike Lewis},
  \bibinfo{person}{Luke Zettlemoyer}, {and} \bibinfo{person}{Veselin
  Stoyanov}.} \bibinfo{year}{2019}\natexlab{}.
\newblock \showarticletitle{Roberta: A robustly optimized bert pretraining
  approach}.
\newblock \bibinfo{journal}{\emph{arXiv e-prints}} (\bibinfo{year}{2019}).
\newblock
\showeprint[arxiv]{1907.11692}


\bibitem[\protect\citeauthoryear{Mao, Xiao, Zhu, Lu, Tang, and He}{Mao
  et~al\mbox{.}}{2020}]%
        {mao2020item}
\bibfield{author}{\bibinfo{person}{Kelong Mao}, \bibinfo{person}{Xi Xiao},
  \bibinfo{person}{Jieming Zhu}, \bibinfo{person}{Biao Lu},
  \bibinfo{person}{Ruiming Tang}, {and} \bibinfo{person}{Xiuqiang He}.}
  \bibinfo{year}{2020}\natexlab{}.
\newblock \showarticletitle{Item Tagging for Information Retrieval: A
  Tripartite Graph Neural Network based Approach}. In
  \bibinfo{booktitle}{\emph{Proceedings of the 43rd International ACM SIGIR
  Conference on Research and Development in Information Retrieval}}.
  \bibinfo{pages}{2327--2336}.
\newblock


\bibitem[\protect\citeauthoryear{McPherson, Smith-Lovin, and Cook}{McPherson
  et~al\mbox{.}}{2001}]%
        {mcpherson2001birds}
\bibfield{author}{\bibinfo{person}{Miller McPherson}, \bibinfo{person}{Lynn
  Smith-Lovin}, {and} \bibinfo{person}{James~M Cook}.}
  \bibinfo{year}{2001}\natexlab{}.
\newblock \showarticletitle{Birds of a feather: Homophily in social networks}.
\newblock \bibinfo{journal}{\emph{Annual review of sociology}}
  \bibinfo{volume}{27}, \bibinfo{number}{1} (\bibinfo{year}{2001}),
  \bibinfo{pages}{415--444}.
\newblock


\bibitem[\protect\citeauthoryear{Mikolov, Sutskever, Chen, Corrado, and
  Dean}{Mikolov et~al\mbox{.}}{2013}]%
        {mikolov2013distributed}
\bibfield{author}{\bibinfo{person}{Tomas Mikolov}, \bibinfo{person}{Ilya
  Sutskever}, \bibinfo{person}{Kai Chen}, \bibinfo{person}{Greg~S Corrado},
  {and} \bibinfo{person}{Jeff Dean}.} \bibinfo{year}{2013}\natexlab{}.
\newblock \showarticletitle{Distributed representations of words and phrases
  and their compositionality}. In \bibinfo{booktitle}{\emph{NeurIPS}}.
  \bibinfo{pages}{3111--3119}.
\newblock


\bibitem[\protect\citeauthoryear{Nan, Qiao, Xiao, Liu, Leng, Zhang, and Lu}{Nan
  et~al\mbox{.}}{2021}]%
        {nan2021interventional}
\bibfield{author}{\bibinfo{person}{Guoshun Nan}, \bibinfo{person}{Rui Qiao},
  \bibinfo{person}{Yao Xiao}, \bibinfo{person}{Jun Liu},
  \bibinfo{person}{Sicong Leng}, \bibinfo{person}{Hao Zhang}, {and}
  \bibinfo{person}{Wei Lu}.} \bibinfo{year}{2021}\natexlab{}.
\newblock \showarticletitle{Interventional Video Grounding with Dual
  Contrastive Learning}. In \bibinfo{booktitle}{\emph{IEEE Conference on
  Computer Vision and Pattern Recognition}}.
\newblock


\bibitem[\protect\citeauthoryear{Niu, Tang, Zhang, Lu, Hua, and Wen}{Niu
  et~al\mbox{.}}{2021}]%
        {niu2020counterfactual}
\bibfield{author}{\bibinfo{person}{Yulei Niu}, \bibinfo{person}{Kaihua Tang},
  \bibinfo{person}{Hanwang Zhang}, \bibinfo{person}{Zhiwu Lu},
  \bibinfo{person}{Xian-Sheng Hua}, {and} \bibinfo{person}{Ji-Rong Wen}.}
  \bibinfo{year}{2021}\natexlab{}.
\newblock \showarticletitle{Counterfactual vqa: A cause-effect look at language
  bias}.
\newblock \bibinfo{journal}{\emph{IEEE Conference on Computer Vision and
  Pattern Recognition}} (\bibinfo{year}{2021}).
\newblock


\bibitem[\protect\citeauthoryear{Park and Neville}{Park and Neville}{2019}]%
        {park2019exploiting}
\bibfield{author}{\bibinfo{person}{Hogun Park} {and} \bibinfo{person}{Jennifer
  Neville}.} \bibinfo{year}{2019}\natexlab{}.
\newblock \showarticletitle{Exploiting interaction links for node
  classification with deep graph neural networks}. In
  \bibinfo{booktitle}{\emph{IJCAI}}. \bibinfo{pages}{3223--3230}.
\newblock


\bibitem[\protect\citeauthoryear{Pearl}{Pearl}{2009}]%
        {pearl2009causality}
\bibfield{author}{\bibinfo{person}{Judea Pearl}.}
  \bibinfo{year}{2009}\natexlab{}.
\newblock \bibinfo{booktitle}{\emph{Causality}}.
\newblock \bibinfo{publisher}{Cambridge university press}.
\newblock


\bibitem[\protect\citeauthoryear{Pearl}{Pearl}{2019}]%
        {pearl2019seven}
\bibfield{author}{\bibinfo{person}{Judea Pearl}.}
  \bibinfo{year}{2019}\natexlab{}.
\newblock \showarticletitle{The seven tools of causal inference, with
  reflections on machine learning}.
\newblock \bibinfo{journal}{\emph{Commun. ACM}} \bibinfo{volume}{62},
  \bibinfo{number}{3} (\bibinfo{year}{2019}), \bibinfo{pages}{54--60}.
\newblock


\bibitem[\protect\citeauthoryear{Pedregosa, Varoquaux, Gramfort, Michel,
  Thirion, Grisel, Blondel, Prettenhofer, Weiss, Dubourg,
  et~al\mbox{.}}{Pedregosa et~al\mbox{.}}{2011}]%
        {pedregosa2011scikit}
\bibfield{author}{\bibinfo{person}{Fabian Pedregosa}, \bibinfo{person}{Ga{\"e}l
  Varoquaux}, \bibinfo{person}{Alexandre Gramfort}, \bibinfo{person}{Vincent
  Michel}, \bibinfo{person}{Bertrand Thirion}, \bibinfo{person}{Olivier
  Grisel}, \bibinfo{person}{Mathieu Blondel}, \bibinfo{person}{Peter
  Prettenhofer}, \bibinfo{person}{Ron Weiss}, \bibinfo{person}{Vincent
  Dubourg}, {et~al\mbox{.}}} \bibinfo{year}{2011}\natexlab{}.
\newblock \showarticletitle{Scikit-learn: Machine learning in Python}.
\newblock \bibinfo{journal}{\emph{JMLR}}  \bibinfo{volume}{12}
  (\bibinfo{year}{2011}), \bibinfo{pages}{2825--2830}.
\newblock


\bibitem[\protect\citeauthoryear{Ren, Liang, Li, Wang, and de~Rijke}{Ren
  et~al\mbox{.}}{2017}]%
        {ren2017social}
\bibfield{author}{\bibinfo{person}{Zhaochun Ren}, \bibinfo{person}{Shangsong
  Liang}, \bibinfo{person}{Piji Li}, \bibinfo{person}{Shuaiqiang Wang}, {and}
  \bibinfo{person}{Maarten de Rijke}.} \bibinfo{year}{2017}\natexlab{}.
\newblock \showarticletitle{Social collaborative viewpoint regression with
  explainable recommendations}. In \bibinfo{booktitle}{\emph{Proceedings of the
  tenth ACM international conference on web search and data mining}}.
  \bibinfo{pages}{485--494}.
\newblock


\bibitem[\protect\citeauthoryear{Rong, Huang, Xu, and Huang}{Rong
  et~al\mbox{.}}{2019}]%
        {rong2019dropedge}
\bibfield{author}{\bibinfo{person}{Yu Rong}, \bibinfo{person}{Wenbing Huang},
  \bibinfo{person}{Tingyang Xu}, {and} \bibinfo{person}{Junzhou Huang}.}
  \bibinfo{year}{2019}\natexlab{}.
\newblock \showarticletitle{Dropedge: Towards deep graph convolutional networks
  on node classification}. In \bibinfo{booktitle}{\emph{ICLR}}.
\newblock


\bibitem[\protect\citeauthoryear{Saito, Ushiku, Harada, and Saenko}{Saito
  et~al\mbox{.}}{2018}]%
        {saito2018adversarial}
\bibfield{author}{\bibinfo{person}{Kuniaki Saito}, \bibinfo{person}{Yoshitaka
  Ushiku}, \bibinfo{person}{Tatsuya Harada}, {and} \bibinfo{person}{Kate
  Saenko}.} \bibinfo{year}{2018}\natexlab{}.
\newblock \showarticletitle{Adversarial Dropout Regularization}. In
  \bibinfo{booktitle}{\emph{International Conference on Learning
  Representations}}.
\newblock


\bibitem[\protect\citeauthoryear{Tang, Huang, and Zhang}{Tang
  et~al\mbox{.}}{2020a}]%
        {tang2020long}
\bibfield{author}{\bibinfo{person}{Kaihua Tang}, \bibinfo{person}{Jianqiang
  Huang}, {and} \bibinfo{person}{Hanwang Zhang}.}
  \bibinfo{year}{2020}\natexlab{a}.
\newblock \showarticletitle{Long-Tailed Classification by Keeping the Good and
  Removing the Bad Momentum Causal Effect}.
\newblock \bibinfo{journal}{\emph{Advances in Neural Information Processing
  Systems}}  \bibinfo{volume}{33} (\bibinfo{year}{2020}).
\newblock


\bibitem[\protect\citeauthoryear{Tang, Yao, Sun, Wang, Tang, Aggarwal, Mitra,
  and Wang}{Tang et~al\mbox{.}}{2020b}]%
        {tang2020investigating}
\bibfield{author}{\bibinfo{person}{Xianfeng Tang}, \bibinfo{person}{Huaxiu
  Yao}, \bibinfo{person}{Yiwei Sun}, \bibinfo{person}{Yiqi Wang},
  \bibinfo{person}{Jiliang Tang}, \bibinfo{person}{Charu Aggarwal},
  \bibinfo{person}{Prasenjit Mitra}, {and} \bibinfo{person}{Suhang Wang}.}
  \bibinfo{year}{2020}\natexlab{b}.
\newblock \showarticletitle{Investigating and Mitigating Degree-Related Biases
  in Graph Convoltuional Networks}. In \bibinfo{booktitle}{\emph{Proceedings of
  the 29th ACM International Conference on Information \& Knowledge
  Management}}. \bibinfo{pages}{1435--1444}.
\newblock


\bibitem[\protect\citeauthoryear{Vaswani, Shazeer, Parmar, Uszkoreit, Jones,
  Gomez, Kaiser, and Polosukhin}{Vaswani et~al\mbox{.}}{2017}]%
        {multi-head-attention}
\bibfield{author}{\bibinfo{person}{Ashish Vaswani}, \bibinfo{person}{Noam
  Shazeer}, \bibinfo{person}{Niki Parmar}, \bibinfo{person}{Jakob Uszkoreit},
  \bibinfo{person}{Llion Jones}, \bibinfo{person}{Aidan~N Gomez},
  \bibinfo{person}{{\L}ukasz Kaiser}, {and} \bibinfo{person}{Illia
  Polosukhin}.} \bibinfo{year}{2017}\natexlab{}.
\newblock \showarticletitle{Attention is all you need}. In
  \bibinfo{booktitle}{\emph{NeurIPS}}. \bibinfo{pages}{5998--6008}.
\newblock


\bibitem[\protect\citeauthoryear{Veli{\v{c}}kovi{\'c}, Cucurull, Casanova,
  Romero, Lio, and Bengio}{Veli{\v{c}}kovi{\'c} et~al\mbox{.}}{2018}]%
        {GAT}
\bibfield{author}{\bibinfo{person}{Petar Veli{\v{c}}kovi{\'c}},
  \bibinfo{person}{Guillem Cucurull}, \bibinfo{person}{Arantxa Casanova},
  \bibinfo{person}{Adriana Romero}, \bibinfo{person}{Pietro Lio}, {and}
  \bibinfo{person}{Yoshua Bengio}.} \bibinfo{year}{2018}\natexlab{}.
\newblock \showarticletitle{Graph attention networks}.
\newblock \bibinfo{journal}{\emph{ICLR}} (\bibinfo{year}{2018}).
\newblock


\bibitem[\protect\citeauthoryear{Veličković, Fedus, Hamilton, Liò, Bengio,
  and Hjelm}{Veličković et~al\mbox{.}}{2019}]%
        {velickovic2018deep}
\bibfield{author}{\bibinfo{person}{Petar Veličković},
  \bibinfo{person}{William Fedus}, \bibinfo{person}{William~L. Hamilton},
  \bibinfo{person}{Pietro Liò}, \bibinfo{person}{Yoshua Bengio}, {and}
  \bibinfo{person}{R~Devon Hjelm}.} \bibinfo{year}{2019}\natexlab{}.
\newblock \showarticletitle{Deep Graph Infomax}. In
  \bibinfo{booktitle}{\emph{ICLR}}.
\newblock


\bibitem[\protect\citeauthoryear{Wang, Ying, Huang, and Leskovec}{Wang
  et~al\mbox{.}}{2019c}]%
        {wang2019improving}
\bibfield{author}{\bibinfo{person}{Guangtao Wang}, \bibinfo{person}{Rex Ying},
  \bibinfo{person}{Jing Huang}, {and} \bibinfo{person}{Jure Leskovec}.}
  \bibinfo{year}{2019}\natexlab{c}.
\newblock \showarticletitle{Improving graph attention networks with large
  margin-based constraints}.
\newblock \bibinfo{journal}{\emph{arXiv preprint arXiv:1910.11945}}
  (\bibinfo{year}{2019}).
\newblock


\bibitem[\protect\citeauthoryear{Wang, Ying, Huang, and Leskovec}{Wang
  et~al\mbox{.}}{2020}]%
        {wang2020direct}
\bibfield{author}{\bibinfo{person}{Guangtao Wang}, \bibinfo{person}{Rex Ying},
  \bibinfo{person}{Jing Huang}, {and} \bibinfo{person}{Jure Leskovec}.}
  \bibinfo{year}{2020}\natexlab{}.
\newblock \showarticletitle{Direct Multi-hop Attention based Graph Neural
  Network}.
\newblock \bibinfo{journal}{\emph{arXiv preprint arXiv:2009.14332}}
  (\bibinfo{year}{2020}).
\newblock


\bibitem[\protect\citeauthoryear{Wang, Feng, He, Zhang, and Chua}{Wang
  et~al\mbox{.}}{2021}]%
        {wang2020click}
\bibfield{author}{\bibinfo{person}{Wenjie Wang}, \bibinfo{person}{Fuli Feng},
  \bibinfo{person}{Xiangnan He}, \bibinfo{person}{Hanwang Zhang}, {and}
  \bibinfo{person}{Tat-Seng Chua}.} \bibinfo{year}{2021}\natexlab{}.
\newblock \showarticletitle{" Click" Is Not Equal to" Like": Counterfactual
  Recommendation for Mitigating Clickbait Issue}.
\newblock \bibinfo{journal}{\emph{Proceedings of the 44th International ACM
  SIGIR Conference on Research and Development in Information Retrieval}}
  (\bibinfo{year}{2021}).
\newblock


\bibitem[\protect\citeauthoryear{Wang, He, Wang, Feng, and Chua}{Wang
  et~al\mbox{.}}{2019a}]%
        {NGCF}
\bibfield{author}{\bibinfo{person}{Xiang Wang}, \bibinfo{person}{Xiangnan He},
  \bibinfo{person}{Meng Wang}, \bibinfo{person}{Fuli Feng}, {and}
  \bibinfo{person}{Tat{-}Seng Chua}.} \bibinfo{year}{2019}\natexlab{a}.
\newblock \showarticletitle{Neural Graph Collaborative Filtering}. In
  \bibinfo{booktitle}{\emph{SIGIR}}. \bibinfo{pages}{165--174}.
\newblock


\bibitem[\protect\citeauthoryear{Wang, Ji, Shi, Wang, Ye, Cui, and Yu}{Wang
  et~al\mbox{.}}{2019b}]%
        {wang2019heterogeneous}
\bibfield{author}{\bibinfo{person}{Xiao Wang}, \bibinfo{person}{Houye Ji},
  \bibinfo{person}{Chuan Shi}, \bibinfo{person}{Bai Wang},
  \bibinfo{person}{Yanfang Ye}, \bibinfo{person}{Peng Cui}, {and}
  \bibinfo{person}{Philip~S Yu}.} \bibinfo{year}{2019}\natexlab{b}.
\newblock \showarticletitle{Heterogeneous graph attention network}. In
  \bibinfo{booktitle}{\emph{WWW}}. \bibinfo{pages}{2022--2032}.
\newblock


\bibitem[\protect\citeauthoryear{Xinyi and Chen}{Xinyi and Chen}{2019}]%
        {xinyi2018capsule}
\bibfield{author}{\bibinfo{person}{Zhang Xinyi} {and} \bibinfo{person}{Lihui
  Chen}.} \bibinfo{year}{2019}\natexlab{}.
\newblock \showarticletitle{Capsule Graph Neural Network}. In
  \bibinfo{booktitle}{\emph{ICLR}}.
\newblock


\bibitem[\protect\citeauthoryear{Xu, Shen, Cao, Qiu, and Cheng}{Xu
  et~al\mbox{.}}{2019c}]%
        {xu2018graph}
\bibfield{author}{\bibinfo{person}{Bingbing Xu}, \bibinfo{person}{Huawei Shen},
  \bibinfo{person}{Qi Cao}, \bibinfo{person}{Yunqi Qiu}, {and}
  \bibinfo{person}{Xueqi Cheng}.} \bibinfo{year}{2019}\natexlab{c}.
\newblock \showarticletitle{Graph Wavelet Neural Network}. In
  \bibinfo{booktitle}{\emph{ICLR}}.
\newblock


\bibitem[\protect\citeauthoryear{Xu, Cui, Hong, Zhang, Yang, and Liu}{Xu
  et~al\mbox{.}}{2019a}]%
        {GIL}
\bibfield{author}{\bibinfo{person}{Chunyan Xu}, \bibinfo{person}{Zhen Cui},
  \bibinfo{person}{Xiaobin Hong}, \bibinfo{person}{Tong Zhang},
  \bibinfo{person}{Jian Yang}, {and} \bibinfo{person}{Wei Liu}.}
  \bibinfo{year}{2019}\natexlab{a}.
\newblock \showarticletitle{Graph inference learning for semi-supervised
  classification}. In \bibinfo{booktitle}{\emph{ICLR}}.
\newblock


\bibitem[\protect\citeauthoryear{Xu, Hu, Leskovec, and Jegelka}{Xu
  et~al\mbox{.}}{2019b}]%
        {xu2018powerful}
\bibfield{author}{\bibinfo{person}{Keyulu Xu}, \bibinfo{person}{Weihua Hu},
  \bibinfo{person}{Jure Leskovec}, {and} \bibinfo{person}{Stefanie Jegelka}.}
  \bibinfo{year}{2019}\natexlab{b}.
\newblock \showarticletitle{How Powerful are Graph Neural Networks?}
\newblock \bibinfo{journal}{\emph{ICLR}} (\bibinfo{year}{2019}).
\newblock


\bibitem[\protect\citeauthoryear{Xu, Li, Tian, Sonobe, Kawarabayashi, and
  Jegelka}{Xu et~al\mbox{.}}{2018}]%
        {JKNet}
\bibfield{author}{\bibinfo{person}{Keyulu Xu}, \bibinfo{person}{Chengtao Li},
  \bibinfo{person}{Yonglong Tian}, \bibinfo{person}{Tomohiro Sonobe},
  \bibinfo{person}{Ken-ichi Kawarabayashi}, {and} \bibinfo{person}{Stefanie
  Jegelka}.} \bibinfo{year}{2018}\natexlab{}.
\newblock \showarticletitle{Representation learning on graphs with jumping
  knowledge networks}.
\newblock \bibinfo{journal}{\emph{ICML}} (\bibinfo{year}{2018}),
  \bibinfo{pages}{8676--8685}.
\newblock


\bibitem[\protect\citeauthoryear{Yang, Feng, Ji, Wang, and Chua}{Yang
  et~al\mbox{.}}{2021}]%
        {Yang2021DVMR}
\bibfield{author}{\bibinfo{person}{Xun Yang}, \bibinfo{person}{Fuli Feng},
  \bibinfo{person}{Wei Ji}, \bibinfo{person}{Meng Wang}, {and}
  \bibinfo{person}{Tat-Seng Chua}.} \bibinfo{year}{2021}\natexlab{}.
\newblock \showarticletitle{Deconfounded Video Moment Retrieval with Causal
  Intervention}. In \bibinfo{booktitle}{\emph{Proceedings of the 44th
  International ACM SIGIR Conference on Research and Development in Information
  Retrieval}}.
\newblock


\bibitem[\protect\citeauthoryear{Yang}{Yang}{2020}]%
        {yang2020biomedical}
\bibfield{author}{\bibinfo{person}{Zuoxi Yang}.}
  \bibinfo{year}{2020}\natexlab{}.
\newblock \showarticletitle{Biomedical Information Retrieval incorporating
  Knowledge Graph for Explainable Precision Medicine}. In
  \bibinfo{booktitle}{\emph{Proceedings of the 43rd International ACM SIGIR
  Conference on Research and Development in Information Retrieval}}.
  \bibinfo{pages}{2486--2486}.
\newblock


\bibitem[\protect\citeauthoryear{Ying, Bourgeois, You, Zitnik, and
  Leskovec}{Ying et~al\mbox{.}}{2019}]%
        {ying2019gnnexplainer}
\bibfield{author}{\bibinfo{person}{Zhitao Ying}, \bibinfo{person}{Dylan
  Bourgeois}, \bibinfo{person}{Jiaxuan You}, \bibinfo{person}{Marinka Zitnik},
  {and} \bibinfo{person}{Jure Leskovec}.} \bibinfo{year}{2019}\natexlab{}.
\newblock \showarticletitle{Gnnexplainer: Generating explanations for graph
  neural networks}. In \bibinfo{booktitle}{\emph{NeurIPS}}.
  \bibinfo{pages}{9244--9255}.
\newblock


\bibitem[\protect\citeauthoryear{Yue, Wang, Zhang, Sun, and Hua}{Yue
  et~al\mbox{.}}{2021}]%
        {yue2021counterfactual}
\bibfield{author}{\bibinfo{person}{Zhongqi Yue}, \bibinfo{person}{Tan Wang},
  \bibinfo{person}{Hanwang Zhang}, \bibinfo{person}{Qianru Sun}, {and}
  \bibinfo{person}{Xian-Sheng Hua}.} \bibinfo{year}{2021}\natexlab{}.
\newblock \showarticletitle{Counterfactual Zero-Shot and Open-Set Visual
  Recognition}.
\newblock \bibinfo{journal}{\emph{IEEE Conference on Computer Vision and
  Pattern Recognition}} (\bibinfo{year}{2021}).
\newblock


\bibitem[\protect\citeauthoryear{Yue, Zhang, Sun, and Hua}{Yue
  et~al\mbox{.}}{2020}]%
        {yue2020interventional}
\bibfield{author}{\bibinfo{person}{Zhongqi Yue}, \bibinfo{person}{Hanwang
  Zhang}, \bibinfo{person}{Qianru Sun}, {and} \bibinfo{person}{Xian-Sheng
  Hua}.} \bibinfo{year}{2020}\natexlab{}.
\newblock \showarticletitle{Interventional Few-Shot Learning}.
\newblock \bibinfo{journal}{\emph{Advances in Neural Information Processing
  Systems}}  \bibinfo{volume}{33} (\bibinfo{year}{2020}).
\newblock


\bibitem[\protect\citeauthoryear{Yun, Jeong, Kim, Kang, and Kim}{Yun
  et~al\mbox{.}}{2019}]%
        {yun2019graph}
\bibfield{author}{\bibinfo{person}{Seongjun Yun}, \bibinfo{person}{Minbyul
  Jeong}, \bibinfo{person}{Raehyun Kim}, \bibinfo{person}{Jaewoo Kang}, {and}
  \bibinfo{person}{Hyunwoo~J Kim}.} \bibinfo{year}{2019}\natexlab{}.
\newblock \showarticletitle{Graph transformer networks}. In
  \bibinfo{booktitle}{\emph{NeurIPS}}. \bibinfo{pages}{11983--11993}.
\newblock


\bibitem[\protect\citeauthoryear{Zhang, Jiang, Wang, Kuang, Zhao, Zhu, Yu,
  Yang, and Wu}{Zhang et~al\mbox{.}}{2020c}]%
        {zhang2020devlbert}
\bibfield{author}{\bibinfo{person}{Shengyu Zhang}, \bibinfo{person}{Tan Jiang},
  \bibinfo{person}{Tan Wang}, \bibinfo{person}{Kun Kuang},
  \bibinfo{person}{Zhou Zhao}, \bibinfo{person}{Jianke Zhu},
  \bibinfo{person}{Jin Yu}, \bibinfo{person}{Hongxia Yang}, {and}
  \bibinfo{person}{Fei Wu}.} \bibinfo{year}{2020}\natexlab{c}.
\newblock \showarticletitle{DeVLBert: Learning Deconfounded Visio-Linguistic
  Representations}. In \bibinfo{booktitle}{\emph{Proceedings of the 28th ACM
  International Conference on Multimedia}}. \bibinfo{pages}{4373--4382}.
\newblock


\bibitem[\protect\citeauthoryear{Zhang, Yao, Zhao, Chua, and Wu}{Zhang
  et~al\mbox{.}}{2021}]%
        {zhang2021cause}
\bibfield{author}{\bibinfo{person}{Shengyu Zhang}, \bibinfo{person}{Dong Yao},
  \bibinfo{person}{Zhou Zhao}, \bibinfo{person}{Tat-Seng Chua}, {and}
  \bibinfo{person}{Fei Wu}.} \bibinfo{year}{2021}\natexlab{}.
\newblock \showarticletitle{CauseRec: Counterfactual User Sequence Synthesis
  for Sequential Recommendation}. In \bibinfo{booktitle}{\emph{Proceedings of
  the 44th International ACM SIGIR Conference on Research and Development in
  Information Retrieval}}.
\newblock


\bibitem[\protect\citeauthoryear{Zhang, Deng, and Lam}{Zhang
  et~al\mbox{.}}{2020b}]%
        {zhang2020answer}
\bibfield{author}{\bibinfo{person}{Wenxuan Zhang}, \bibinfo{person}{Yang Deng},
  {and} \bibinfo{person}{Wai Lam}.} \bibinfo{year}{2020}\natexlab{b}.
\newblock \showarticletitle{Answer ranking for product-related questions via
  multiple semantic relations modeling}. In
  \bibinfo{booktitle}{\emph{Proceedings of the 43rd International ACM SIGIR
  Conference on Research and Development in Information Retrieval}}.
  \bibinfo{pages}{569--578}.
\newblock


\bibitem[\protect\citeauthoryear{Zhang, Cui, and Zhu}{Zhang
  et~al\mbox{.}}{2020a}]%
        {zhang2018deep}
\bibfield{author}{\bibinfo{person}{Ziwei Zhang}, \bibinfo{person}{Peng Cui},
  {and} \bibinfo{person}{Wenwu Zhu}.} \bibinfo{year}{2020}\natexlab{a}.
\newblock \showarticletitle{Deep learning on graphs: A survey}.
\newblock \bibinfo{journal}{\emph{TKDE}} (\bibinfo{year}{2020}).
\newblock


\bibitem[\protect\citeauthoryear{Zhou}{Zhou}{2012}]%
        {zhou2012ensemble}
\bibfield{author}{\bibinfo{person}{Zhi-Hua Zhou}.}
  \bibinfo{year}{2012}\natexlab{}.
\newblock \bibinfo{booktitle}{\emph{Ensemble methods: foundations and
  algorithms}}.
\newblock \bibinfo{publisher}{Chapman and Hall/CRC}.
\newblock


\end{thebibliography}
